\documentclass[runningheads]{llncs}
\usepackage{eccv}
\usepackage[accsupp]{axessibility}
\usepackage{array}
\usepackage{booktabs}
\usepackage{comment}
\usepackage{eccvabbrv}
\usepackage{graphicx}
\usepackage{hyperref}
\usepackage{listings}
\lstdefinelanguage{json}{
  showstringspaces=false,
  breaklines=true,
  morestring=[b]",
}
\usepackage{orcidlink}

\newcommand{\method}{\textsc{SynCity~3000}\xspace}

\makeatletter
\renewcommand{\paragraph}{%
  \@startsection{paragraph}{4}%
  {\z@}{-0.5em}{-0.5em}%
  {\normalfont\normalsize\bfseries}%
}
\makeatother

\newcommand{\bepsilon}{\boldsymbol{\epsilon}}
\newcommand{\bo}{\boldsymbol{o}}
\newcommand{\bp}{\boldsymbol{p}}
\newcommand{\z}{\boldsymbol{z}}

\title{SynCity 3000\\Bootstrapping Scene-Scale 3D Diffusion}
\titlerunning{SynCity 3000}
\author{Paul Engstler\orcidlink{0009-0009-9725-7497} \and
Iro Laina\orcidlink{0000-0001-8857-7709} \and
Christian Rupprecht\orcidlink{0000-0003-3994-8045} \and
Andrea Vedaldi\orcidlink{0000-0003-1374-2858}}
\authorrunning{P.~Engstler et al.}
\institute{Visual Geometry Group, University of Oxford\\
\email{\tt\small \{paule,iro,chrisr,vedaldi\}@robots.ox.ac.uk}\\
\url{https://research.paulengstler.com/syncity-3k}
}

\begin{document}
\maketitle

\begin{abstract}
We present \method, a framework for generating 3D scenes that are globally coherent while enabling fine-grained layout control.
Building on the ability of current image-to-3D generators to produce complex 3D assets from a single image, we extend this capability to the scale of entire scenes by adapting the generator to be applicable as a convolutional operator.
We achieve this by fine-tuning the model on scene-like data generated by a new synthetic data engine, which we propose to address the scarcity of 3D scene data for training.
The convolutional generator is then applied to a dimetric image of the entire scene, generated from the user prompt, resulting in 3D scenes of arbitrary size and complexity.
Across diverse prompts and layouts, \method produces large, coherent, and detailed scenes, addressing the shortcomings of prior approaches to 3D scene generation.
\end{abstract}
\section{Introduction}%
\label{sec:intro}

Creating 3D content for movies, games and simulations is a time-consuming and labor-intensive task that requires skilled artists and designers.
Recent models that can produce automatically high-quality 3D assets from text prompts~\cite{xiang2024structured,zhao2025hunyuan3d,hunyuan3d2025hunyuan3d,li2025triposg}, but these are usually limited to single objects.
Generating entire 3D scenes would be much more impactful in applications.
SynCity~\cite{engstler2025syncity} has recently demonstrated that off-the-shelf models for image generation and image-based 3D reconstruction of single objects can be repurposed to generate large scenes.
By building on off-the-shelf models, SynCity sidesteps the lack of large and diverse 3D scene datasets for training a corresponding generator from scratch.
It achieves this by reinterpreting the scene as a grid of tiles, each of which is akin to an object which can be generated semi-independently.
While this works, the grid-like structure is clearly visible in the final output.

\begin{figure}[t]
\centering
\includegraphics[width=\linewidth]{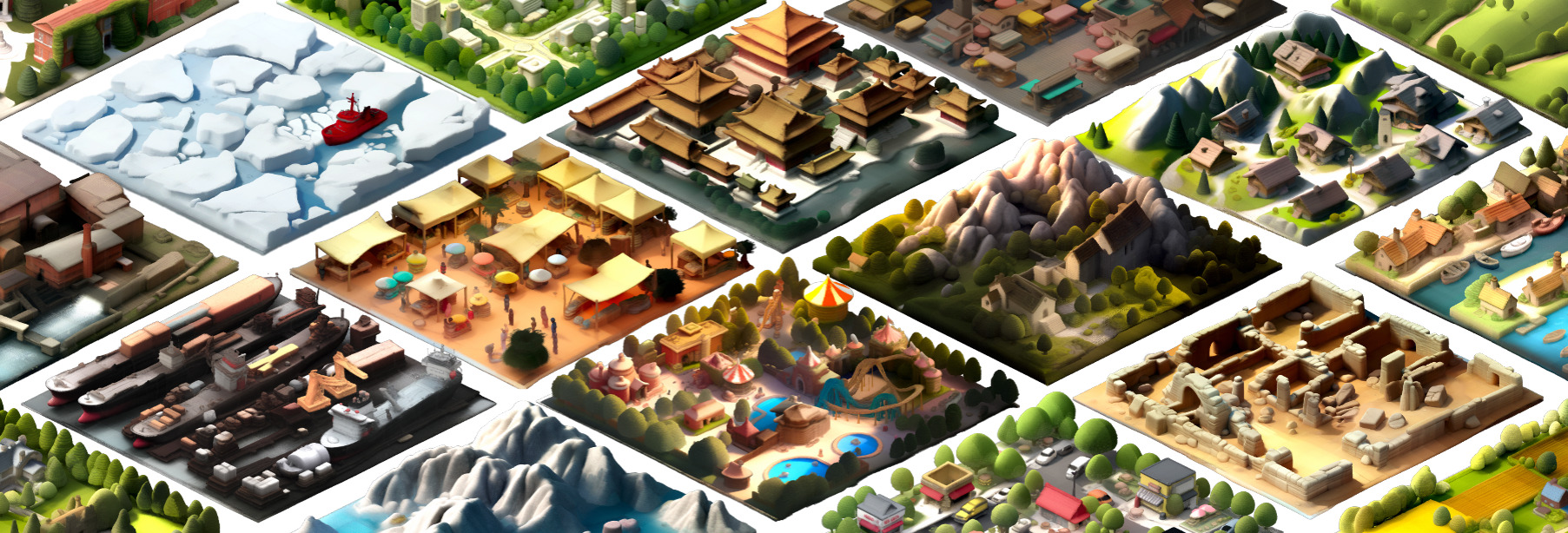}
\caption{Diverse 3D worlds of arbitrary size and complexity are easily created with \method from scratch.
Our approach first constructs a visually and semantically coherent 2D template of the entire scene, and then converts it into 3D Gaussian Splats with a fine-tuned two-stage generative diffusion model.}%
\label{fig:splash}
\vspace{-1em}
\end{figure}

In this paper, we address this limitation by introducing \method, a two-stage framework for generating large-scale 3D scenes from text prompts (\cref{fig:splash,fig:method}).
Breaking free from the fixed grid-like structure of SynCity, our approach creates 3D worlds of arbitrary structure and complexity.
In the first stage, we generate a 2D image template that defines the appearance and layout of the scene.
In the second stage, this template is converted into the final 3D scene.
This pipeline operates automatically without manual intervention.
Notably, the two stages are independent, allowing any template-like image, including those created by graphic artists, to be used for generating a 3D scene.

The stages require repurposing off-the-shelf models for 2D and 3D generation.
This necessitates several innovations, which we summarize next and in \cref{sec:method}.

In the first stage, we prompt an image generator to create a high-resolution 2D template of the scene.
The generator is based on latent diffusion and, in order to generate scenes of any extent, we divide the 2D latent space into partially overlapping windows without forcing them to look like square tiles.
Optional layout constraints can be introduced to allow fine-grained control of the generated scene.
By conditioning on all constraints and averaging their contributions at each step of the image diffusion process (inspired by MultiDiffusion~\cite{bar2023multidiffusion}), windows blend in naturally with their surroundings.

In the second stage, we use an image-conditioned 3D generation model to transform the template into a 3D scene.
We employ a voxel-based approach, where a coarse, voxelized version of the world is generated first and then enriched with rich visual and semantic features for each voxel.
These are then decoded into highly detailed scenes represented by 3D Gaussian Splats.
Similar to the template generation, we subdivide the scene (or the corresponding latent code) into partially overlapping windows to support scenes of arbitrary size.
We fine-tune the 3D generator to operate on the sliding window in a ``convolutional'' manner, and also introduce additional adaptation that facilitates learning this new behaviour.
Fine-tuning uses synthetic data we create procedurally for this purpose, as detailed in \cref{sec:dataset}.

Finally, in \cref{sec:experiments}, we demonstrate the effectiveness of \method on a variety of scenes of different sizes and complexities.
Empirically, we show that our approach outperforms SynCity and other reconstruction methods both qualitatively and quantitatively (\cref{fig:syncity-comparison,fig:recon-comparison,tab:perceptual}).
\section{Related work}%
\label{sec:related-work}

\paragraph{Image-based scene generation.}

The ability of 2D generative models to generate increasingly realistic images while exhibiting a semantic and geometric understanding of the depicted scene~\cite{zhan2024general,chen2023beyond} has inspired a wide range of works~\cite{hollein2023text2room,zhang23text2nerf:,ouyang23text2immersion:,chung23luciddreamer:,engstler25a,shriram2024realmdreamer,yu2024wonderjourney,yu2024wonderworld,wang23prolificdreamer:,Po2023Compositional3S} that utilize them to create representations of large-scale scenes in the form of meshes, Neural Radiance Fields~\cite{mildenhall2021nerf}, or 3D Gaussian Splats~\cite{kerbl20233d}.
Most of these methods leverage monocular depth estimation models~\cite{Bae2022,bhat2023zoedepth,ke2024repurposing,yang2024depth} to project images and apply various heuristics to fuse them in space.
Other works directly operate on panoramas~\cite{stan2023ldm3d,perf2023,li24scenedreamer360:,wu2023panodiffusion} or posed images~\cite{kim2023nfldm}.
However, the generated scenes frequently exhibit geometric artifacts and holes in the scene due to occlusions when steering away from the viewpoints used to generate the scene.

\paragraph{Asset-based scene generation.}

By populating a scene with 3D assets based on a synthesized layout, geometric and occlusion issues are easily resolved.
In this line of work, layouts are either directly based on an image~\cite{dai24automated,gao24diffcad:,ardelean2024gen3dsr,huang2025midi,yao2025cast,gumeli22roca:,kuo21patch2cad:}, or generated with large language models~\cite{feng2023layoutgpt,hu24scenecraft:,zhou2024gala3d,yang24holodeck:,ocal24sceneteller:,sun25layoutvlm:} or trained diffusion models~\cite{tang24diffuscene:, lin24instructscene:, zhai24echoscene:}.
Assets may be retrieved from a library or generated on-the-fly, enabled by object-centric 3D generation models~\cite{xiang2024structured,zhao25hunyuan3d,li25triposg:,li2025sparc3d}.
The generated scenes, however, are frequently monotonous, at times unrealistic, suffer from inter-object collisions, and are largely restricted to indoor scenes.

\paragraph{Explicit scene generation.}

Geometric consistency issues can be circumvented by directly synthesizing the scene representation, as presented in more recent works.
LT3SD~\cite{meng2025lt3sd} proposes to learn a diffusion model, which employs a patch-by-patch and coarse-to-fine approach to build 3D environments without conditioning.
$\mathcal{X}^3$~\cite{ren2024xcube} can generate large-scale 3D scenes by representing them as a hierarchy of sparse voxel grids.
Both require large-scale, domain-specific training data, and are either incapable or struggle with the conditioning of their generation process.
BlockFusion~\cite{wu24blockfusion:} learns a diffusion model to extend a mesh auto-regressively by small blocks, SceneCraft~\cite{yang2024scenecraft} one to generate scene renderings.
While the layout of the generated scene can be controlled, these methods require the availability of domain-specific datasets.
Recent work~\cite{lee2023diffusion,lee2024semcity} generates large-scale scenes but focuses on semantic categories and thus on semantic completion of an existing scene rather than providing text or image interfaces.

In SynCity~\cite{engstler2025syncity}, a scene is reinterpreted as a grid of tiles that are generated individually and then fused together.
This process is steered by prompts on a global and tile level.
3D tiles are generated sequentially in a training-free approach with TRELLIS~\cite{xiang2024structured}, leveraging its ability to create high-quality 3D objects from large-scale 3D object-centric datasets~\cite{deitke23objaverse-xl:, collins22abo:, fu20213d, khanna2024habitat}.
While the generated scenes follow an overall theme, the sequential tile generation pattern causes scenes to lack overall coherence, leading to a busy, grid-like appearance.

Another training-free approach based on TRELLIS~\cite{xiang2024structured}, 3DTown~\cite{zheng25constructing}, transforms an image of a scene into 3D Gaussian Splats.
Here, a monocular depth estimator~\cite{wang25vggt} is used to build a geometric prior of the scene.
It is then partitioned into overlapping regions, locking parts already observed through the image.
The unknown parts are then completed by adding a mask in the 3D generation process.
While the generated scenes follow the input image, they suffer from geometric artifacts, holes, and limited resolution.

In NuiScene~\cite{lee25nuiscene:}, a conditional outpainting diffusion model is trained on a subset of Objaverse~\cite{deitke23objaverse-xl:}, which contains 43 scenes.
Scenes are produced in chunks, based on a learned VecSet~\cite{3DShape2VecSet} encoding, enabling unbounded generation.
However, there are visible discontinuities between chunks, textures have to be generated with SceneTex~\cite{chen24scenetex:}, and the small dataset size severely limits the method's ability to generalize.

In this work, we address the limitations of these prior works.
We propose a simple method inspired by MultiDiffusion~\cite{bar-tal23multidiffusion:} (and similar to GLIGEN~\cite{li23gligen:}) to create an arbitrarily complex scene template, which strictly generalizes SynCity.
To be able to turn this template into a 3D representation, we fine-tune a 3D generation model~\cite{xiang2024structured} to support convolutional inference, thus tailoring it to the task rather than building brittle heuristics to use an object-centric model, as in 3DTown.
This fine-tuning is enabled by our proposed synthetic dataset engine, which generates scene-like data, allowing us to evade the lack of large-scale 3D scene datasets that are not strictly limited to indoor settings~\cite{dai17scannet:, armeni20193d, yeshwanth2023scannet, ramakrishnan2021habitat,pan2023aria,khanna2024habitat,roberts2021hypersim}.
This engine allows us to scale far beyond the training dataset of NuiScene.

\begin{figure*}[t]
\centering
\includegraphics[width=\textwidth]{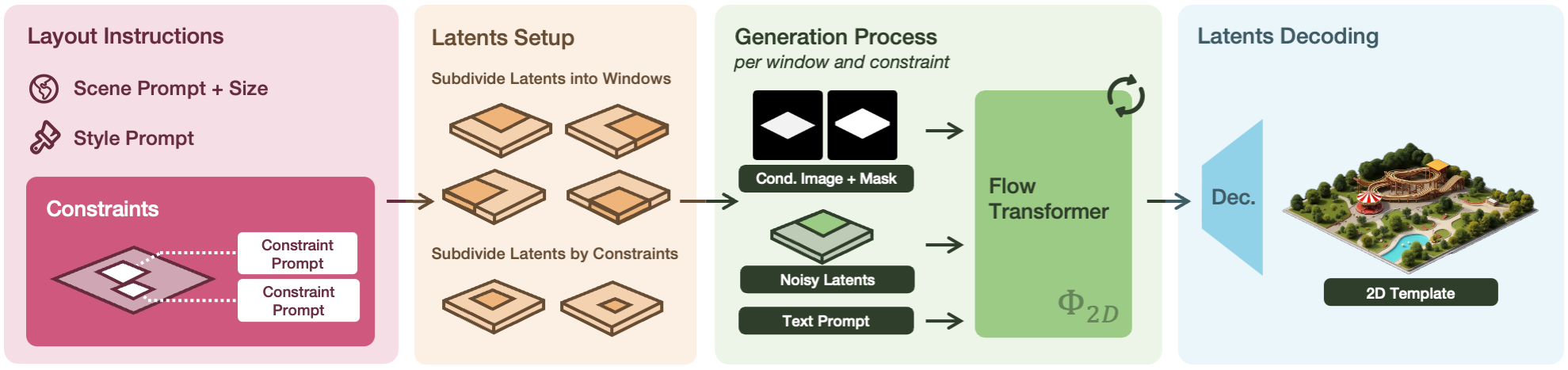}
\caption{\textbf{2D Generation Pipeline.} We generate a dimetric scene template based on text prompts and optional layout constraints. After subdividing the latents into windows, we apply a latent diffusion model to jointly denoise the latent canvas. Once complete, the latent canvas is then decoded into the final 2D scene template.}%
\label{fig:2d-gen}
\end{figure*}

\section{Method}%
\label{sec:method}

The input to \method is a high-level textual description of the scene which consists of the theme prompt $P$, which describes the overall theme of the scene (e.g., ``a bustling medieval town square''), and the style prompt $P_\text{style}$, which describes its visual appearance (e.g., ``a bright sunny day'').
Optionally, the method can also take as input a set of \emph{local prompts} $P_w$ (e.g., ``a stone well'') each associated with a \emph{window} $w \in \mathcal{W}_c$, further defined below.

The output of our method is a full 3D scene represented as 3D Gaussian Splats~\cite{xiang2024structured}, which can be rendered from arbitrary viewpoints.

As illustrated in \cref{fig:2d-gen} and \cref{fig:method}, \method{} comprises two independent stages:
the first stage (\cref{sec:2d-template}) generates a 2D template of the scene and the second stage (\cref{sec:3d-generation}) generates a corresponding 3D model.

\subsection{2D template generation}%
\label{sec:2d-template}

Stage~1 generates a 2D image \emph{template} $J$ that defines the visual appearance and layout of the scene as a whole.
As in SynCity, this stage repurposes a pre-trained text-to-image model $\Phi_\text{2D}$ that uses latent denoising diffusion.
Although off-the-shelf, the model does not directly generate images suitable for 3D reconstruction.
We develop a prompting strategy to enable the model to generate \emph{dimetric views}.
We follow SynCity and run the generator in ``inpainting mode'', passing to it a ``base image'' $I_\text{base}$ that shows a dimetric perspective of a supporting structure (similar to a concrete slab), and an inpainting mask $M_\text{base}$ that covers the area above this base, as shown in \cref{fig:2d-gen}.
The inpainted image output by $\Phi_\text{2D}$ then has the required dimetric framing.

A limitation of the scheme above is that the extent of the scene is bound by the maximum resolution of the generator $\Phi_\text{2D}$\@.
SynCity addressed this issue by generating and combining non-overlapping scene tiles, one at a time.
While this process can be extended indefinitely, it is prone to visual artifacts when tiles are reconstructed in 3D and assembled.
To address this issue, we propose a new approach that results in a much better and more coherent global structure of the scene.
Inspired by MultiDiffusion~\cite{bar2023multidiffusion}, we apply $\Phi_\text{2D}$ to overlapping windows \emph{throughout} the denoising process.
Thus, information is passed across windows at every denoising step, resulting in a globally coherent image.

In more detail, the denoising model $\Phi_\text{2D}$ operates in \emph{latent space}, given by a grid of tokens
$
\z \in \mathbb{R}^{C \times H/\sigma \times H/\sigma}
$
where $H \times H$ is the resolution of the corresponding 2D image, $\sigma$ is the token downsampling factor, and $C$ is the latent channel dimension.
We assume for simplicity that all images are square, and we call $\z$ the \emph{latent canvas}.

We further assume that the model $\Phi_\text{2D}$ operates on smaller token grids
$
\z_w \in \mathbb{R}^{C \times S/\sigma \times S/\sigma}
$,
where $S \leq H$ is the size of the image window that can be generated by the off-the-shelf model.
Hence, the network $\Phi_\text{2D}$ can be seen as a function
$
\bepsilon_w = \Phi_\text{2D}(\z_w | I_\text{base}, M_\text{base}, P_w, P_\text{style})
$
estimating the noise $\bepsilon_w$ in the code $\z_w$.
Each application of $\Phi_\text{2D}$ is conditioned by the global style prompt $P_\text{style}$ as well as the local content prompt $P_w$, which is window-dependent.
In particular, $P_w = P$ if $w \notin \mathcal{W}_c$, i.e., if the window is not associated with a local prompt.

The windows $\mathcal{W}$ are thus squares of size $S$ sliding over the generated image with a stride $s < S$ to ensure overlap, with the addition of any windows defined by layout constraints $\mathcal{W}_c$ (so that $\mathcal{W}_c\subset\mathcal{W})$.
The symbols $\z_w$ denote the corresponding subset of the latent canvas $\z$ covered by the window $w$.
At each denoising step, the \emph{entire} canvas $\z$ is updated after averaging the estimated noise $\bepsilon_w$ over all windows $w \in \mathcal{W}$, similar to MultiDiffusion~\cite{bar2023multidiffusion}.

\subsection{3D generation}%
\label{sec:3d-generation}

We now turn to the task of transforming the 2D template $J$ obtained in stage~1 (\cref{sec:2d-template}) into a 3D scene.
Similar to 2D generation, SynCity demonstrated that off-the-shelf image-to-3D models like TRELLIS~\cite{xiang2024structured} can be repurposed for this task.
In SynCity, the idea is to reconstruct the 3D scene tile-by-tile.
This works because individual tiles have a complexity that roughly matches a single 3D object and can be reconstructed relatively well by TRELLIS without any further training.
However, there are artifacts where the resulting 3D tiles join up.
Furthermore, it requires tiles to be represented as separated images, which leads to visual and semantic discontinuities as discussed earlier.

As our stage~1 outputs a single, large image of the complete scene in its entirety, we fine-tune TRELLIS to operate in a \emph{convolutional} manner.
This allows us to process the entire scene at once, potentially for an arbitrarily large size of the template image $J$.

TRELLIS itself comprises two generators: the first determines the \emph{sparse structure} of the object, which is a coarse voxel grid outlining its geometry, and the second samples \emph{structured latents}, which are features attached to each voxel that encode shape and appearance details.
Finally, the structured latents are decoded into 3D Gaussian Splats.
We begin by discussing how the first generator can be made convolutional (\cref{sec:convolutional-sparse-structure-gen}), and then we move to the second one (\cref{sec:convolutional-structured-latents-gen}).

\begin{figure*}[t]
\centering
\includegraphics[width=\textwidth]{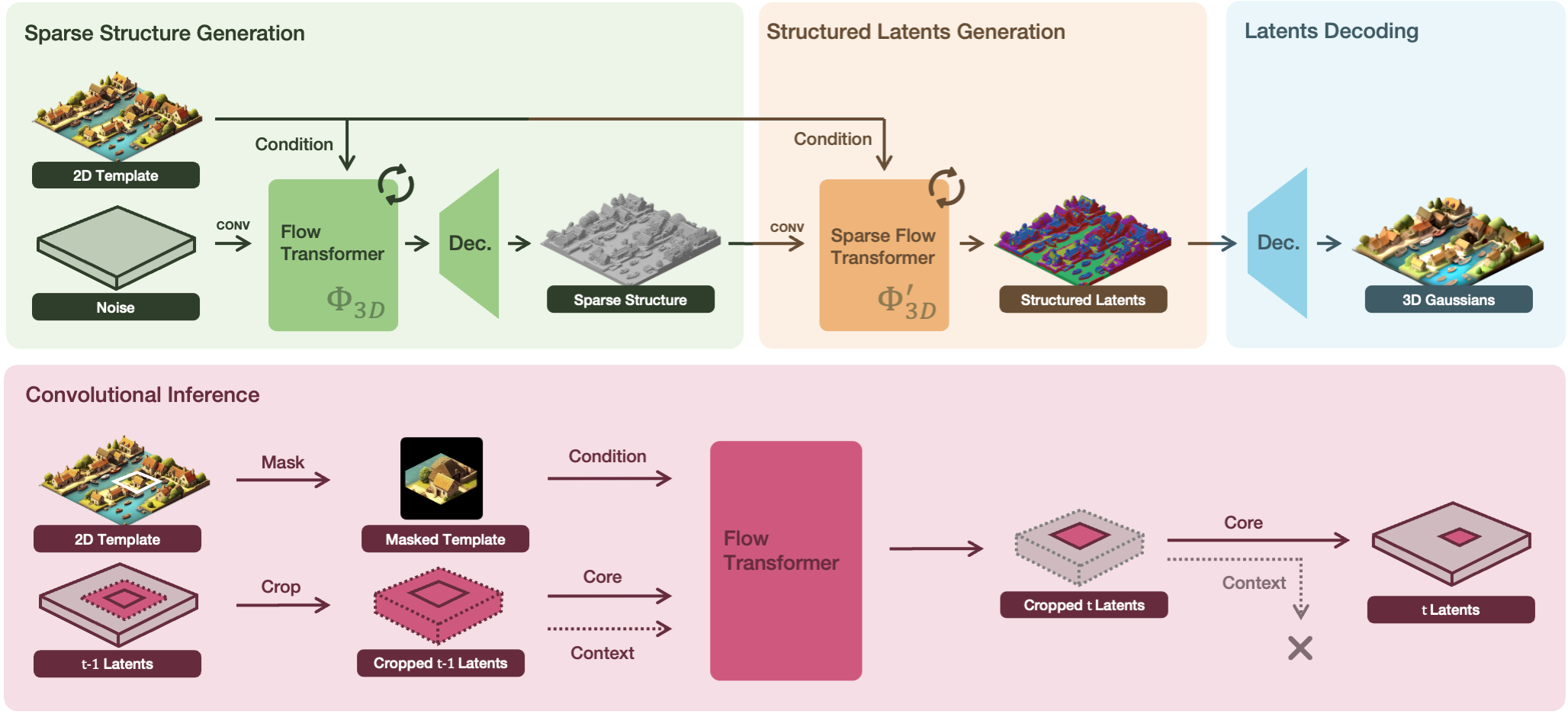}
\caption{\textbf{3D Generation Pipeline.}
We use a multi-stage pipeline to synthesize full 3D scenes from 2D templates using convolutional inference.
In this inference process, we divide the latent into smaller regions that we jointly denoise.
Each region has a direct template correspondence.
To transform a scene, we first synthesize the sparse structure---a coarse voxel grid.
Then, we infer features that encode the appearance and semantics of each voxel.
Finally, we decode these structured latents into the final 3D Gaussian Splats representation.
We use the decoders of~\cite{xiang2024structured}.}%
\label{fig:method}
\end{figure*}

\begin{figure}[t]
\centering
\includegraphics[width=0.65\linewidth]{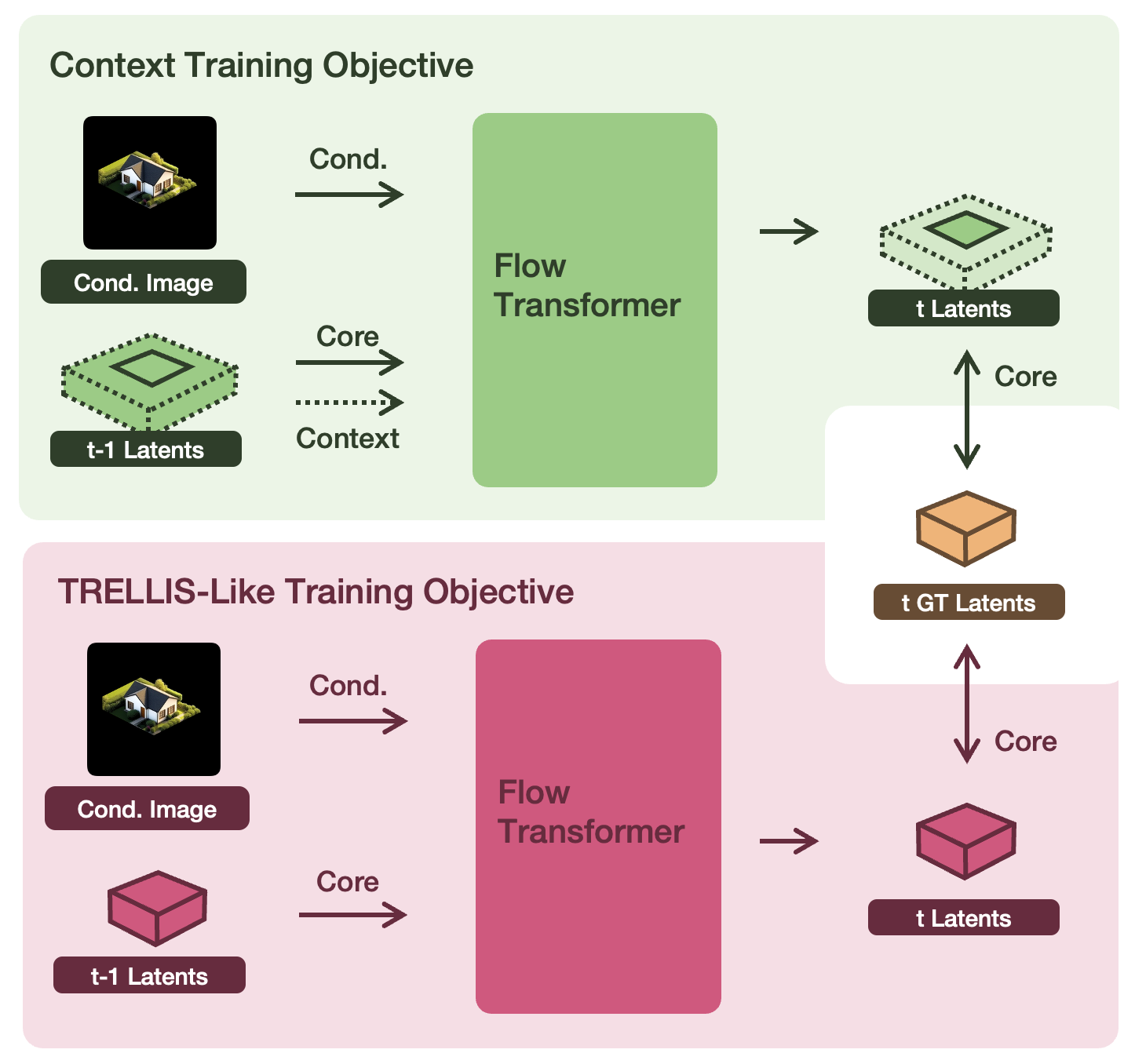}
\caption{\textbf{Fine-Tuning Design.} We utilize two objectives in our fine-tuning process.
In the first (top), we provide the core \textit{and} surrounding context to the model.
In the second (bottom), we only provide the core, which imitates the original objective of TRELLIS~\cite{xiang2024structured}.
In both cases, we use a masked conditioning image and apply the loss only to the core.}%
\label{fig:ft-scheme}
\end{figure}

\subsubsection{Convolutional sparse structure generation.}%
\label{sec:convolutional-sparse-structure-gen}

Given an image $I$ of the object, the first of the two generators in TRELLIS is tasked with sampling a corresponding voxel grid $\mathcal{O} \in \{0, 1\}^{N\times N\times N}$ which captures its rough shape.
TRELLIS auto-encodes $\mathcal{O}$ to a lower-resolution latent grid $\bo \in \mathbb{R}^{M \times M \times M \times C}$, where $M \ll N$ and $C$ is the latent channel dimension.

To reconstruct $\bo$ from an image $I$, TRELLIS uses a denoising Diffusion Transformer (DiT) $\Phi_\text{3D}$.
Let $\bp \in \{0, \dots, M-1\}^{M \times M \times M \times 3}$ be the positional indices of the latent vectors in $\bo$.
The latent grid $\bo$ is reinterpreted as a tensor $M^3 \times C$ of $M^3$ tokens, tokens are summed with sinusoidal positional encodings $\rho(\bp) \in \mathbb{R}^{M^3\times C}$ representing their position in the grid, and the tokens and image are fed to the transformer to obtain the denoising signal $\bepsilon=\Phi_\text{3D}(\bo + \rho(\bp)|I)$.

Our goal is to modify this scheme to reconstruct much wider scenes, i.e., $\mathcal{O} \in \{0, 1\}^{N_\text{scene}\times N_\text{scene}\times N}$ with $N_\text{scene} \gg N$.
The auto-encoder is already convolutional, so, applied to a larger occupancy grid, it produces a correspondingly larger latent grid $\bo \in \mathbb{R}^{M_\text{scene} \times M_\text{scene} \times M \times C}$ with $M_\text{scene} \gg M$.
The denoiser $\Phi_\text{3D}$, however, is not convolutional, so it needs to be modified to be applicable to the new setting.

To apply $\Phi_\text{3D}$ to this larger latent grid, we consider overlapping sliding windows $w \in \mathcal{W}$.
Each window extracts a sub-grid $\bo_w$ of size $M \times M\times M$ from the larger latent grid $\bo$, making it possible to apply the model.
The conditioning signal must also be modified to indicate which portion of the world is encoded by $\bo_w$ and thus needs to be reconstructed.
We do so by considering a corresponding image crop $I_w$ centered on the same portion of the world as $\bo_w$.
Thus, the denoising model is $\bepsilon_w = \Phi_\text{3D}(\bo_w + \rho(\bp) | I_w)$, where the positional encoding $\bp$ remains unchanged, as it is relative to the window rather than the world.
All windows are processed in parallel at each denoising step, and the resulting denoising signals $\bepsilon_w$ are averaged to update the entire latent grid $\bo$.

Note that the statistics of $(\bo,I)$ in the original model are fairly different from those of $(\bo_w, I_w)$ in the convolutional setting above.
For one thing, $\bo_w$ only encodes a well-defined portion of the scene, which we call the \emph{core}, but the image $I_w$ shows not only this portion, but also the area around it, which should not be encoded by $\bo_w$.
We do three things to bridge this gap.

First, as further detailed below, we fine-tune the model $\Phi_\text{3D}$ to operate in this new setting, utilizing the training data we develop in \cref{sec:dataset}.

Second, we focus the model on the core by masking out the part of the image $I_w$ that \emph{does not} correspond to it, passing to it $I_w \odot M$, where $M$ is the mask.
See \cref{fig:method} for an illustration.

Thirdly, we provide further 3D context to the model by expanding the latent grid $\bo_w$ to include voxels surrounding the core, which results in additional contextual tokens $\bo_w^c$ that, in combination with the core, cover an area $M_\text{context}$ larger than the core alone (so overall $M < M_\text{context} \ll M_\text{scene}$).
This helps each application of the model to update the core while accounting for the 3D geometry of the surrounding area.
This also requires assigning positional encodings $\bp^c$ to the context tokens, which we do carefully as explained below.
Hence, the denoising model becomes
$$
\bepsilon_w = \Phi_\text{3D}(\bo_w + \rho(\bp), \bo_w^c + \rho(\bp^c) | I_w \odot M).
$$

\paragraph{Positional encoding for the context tokens.}

Before, we noted that the core tokens $\bo_w$ maintain the same positional encoding $\rho(\bp)$, which are relative to the window, as in the original model.
When no context tokens are provided, this allows the model to operate in a similar manner as the original one as much as possible.
However, when context tokens $\bo_w^c$ are provided, we must assign positional encodings $\rho(\bp^c)$ to them too.
The natural choice is to define them as positions outside the core.

In other words, given the original positional embedding function
$
\rho : \{0,\dots,M-1\}^3 \mapsto [-1, 1]^C
$,
we extend it to the larger domain
$
\{-V, \dots, M + V - 1\}^3
$
where $M_\text{context} = M + 2V$ is the size of the core plus context.
By constructing the grid such that $\bp^c$ span from $(-V, -V, 0)$ to $(M + V - 1, M + V - 1, M - 1)$, the core receives the original positional embeddings.

\paragraph{Fine-tuning scheme.}

Besides the modifications above, we fine-tune TRELLIS to teach it to operate convolutionally.
However, we also wish to keep its ability to generate high-quality 3D objects and prevent catastrophic forgetting.

Thus, we use two separate training objectives to achieve both goals.
Their occurrence is determined by a predefined probability $p$.
We either provide the core \emph{with} or \emph{without} context.
Thanks to our design choices, the latter then mirrors the original task of TRELLIS (see \cref{fig:ft-scheme}).
The conditioning image is the same in both cases.
We use a mean-squared error loss that is only applied to the core.
This reinforces the notion that the models need to consider the context, but only responses within the core are relevant.

\subsubsection{Convolutional structured latents generation.}%
\label{sec:convolutional-structured-latents-gen}

Once the sparse structure $\mathcal{O}$ has been generated, TRELLIS associates to each non-zero voxel at a given position $\bp_i$ a corresponding feature vector $\z_i$ encoding local appearance and shape.
The resulting set of
$
L = \|\mathcal{O}\|_1
$
\emph{structured latents} is stacked in a tensor $\z \in \mathbb{R}^{L \times C}$, with corresponding positional indices $\bp \in \{0,\dots,N-1\}^{L\times 3}$.
These are then passed to one more denoising diffusion model
$\bepsilon = \Phi_\text{3D}'(\z + \rho(\bp) | I)$ to sample $\z$ given the image $I$.
Except for the fact that the tokens $\z$ are sparse instead of dense, this model is very similar to the one described in \cref{sec:convolutional-sparse-structure-gen}, so we can apply the same convolutional scheme.

\subsubsection{Scene inference.}%
\label{sec:scene-inference}

With the models in place, we can now directly use them to transform a template into a complete scene represented by 3D Gaussian Splats, as illustrated in \cref{fig:method}.

Similar to how we set the size of the canvas during the 2D template generation (see \cref{sec:2d-template}), we also set a latent resolution for the 3D generation $\mathcal{R}_\text{latents}$ in integer multiples of the core resolution (i.e., $\mathcal{R}_\text{latents} \mod N = 0$).
This establishes a clear correspondence between pixel (template) and latent space.
Crucially, we are free to choose these values independently, and thus directly control the size and level of detail of the generated scene.

Both the sparse structure and structured latents generation rely on convolutional inference that we have enabled through our design choices and training scheme.
We divide the template and latents into windows using strides $s_\text{pixel}$ and $s_\text{latents}$.
Then, we isolate that particular window by masking the template and cropping the latents.
The latents contain the core, which corresponds to the part of the world shown in the masked template, and its surrounding context.
While the models denoise the latents of the core and context, we only retain its predictions for the core.
At each denoising timestep, we denoise all windows, averaging their contributions if a small stride results in overlaps.

Finally, once we have obtained the complete scene as Gaussians, we apply a simple color correction to address a shortcoming of the TRELLIS model, which we detail in the supplementary materials.
\section{Dataset}%
\label{sec:dataset}

\begin{figure*}[t]
\centering
\includegraphics[width=\textwidth]{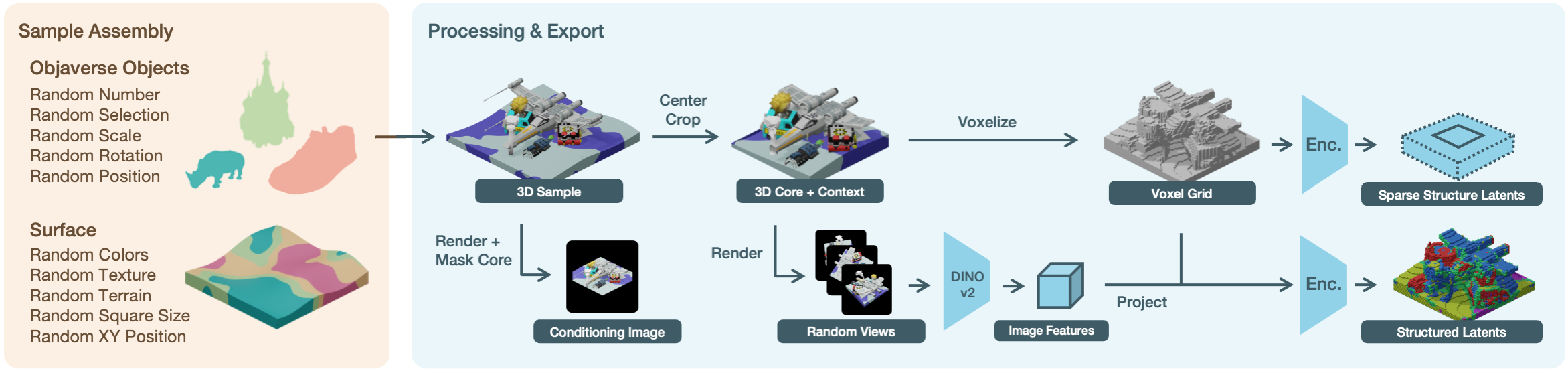}
\caption{\textbf{Data Generation.}
We propose a dataset engine to generate diverse 3D scenes by placing random Objaverse-XL~\cite{deitke23objaverse-xl:} objects on randomized terrains.
Each scene is rendered in dimetric projection to match the 2D template generation~\cref{fig:2d-gen}.
The rendering of the scene is masked to only show the core, whereas we crop the 3D scene to include the core and some context.
This 3D crop is then converted into a voxel grid and infused with DINOv2~\cite{oquab2023dinov2} features.
We then encode it into sparse structure and structured latents using the encoders of~\cite{xiang2024structured}.}%
\label{fig:dataset}
\end{figure*}

While a growing body of large-scale 3D datasets for objects~\cite{deitke23objaverse-xl:, collins22abo:, fu20213d, khanna2024habitat} exists, the collection of scene-scale data has been lagging behind.
Data availability has been a key factor in improving the quality of 3D generative models~\cite{zhang24clay:, xiang2024structured, zhao25hunyuan3d}.
As we aim to fine-tune the models of TRELLIS to operate on large-scale scenes, the lack of 3D scene data poses a problem.

Therefore, we introduce a synthetic dataset engine, which generates complete 3D scenes in a manner that harmonizes with our scene-scale diffusion scheme, as in \cref{sec:3d-generation}.
In the spirit of LRM-Zero~\cite{xie24lrm-zero:}, which uses random assortments of shapes and textures to bootstrap a 3D generator, we place Objaverse-XL objects onto randomly generated surfaces.
For an overview of our proposed generation process, see \cref{fig:dataset}.

\paragraph{Sample assembly.}

To mirror the way we generate our 2D templates (see \cref{sec:2d-template}), we first create a surface of variable size with random terrain.
We blend multiple random colors to texture it, with hard-stop transitions between colors.
Then, we pick a random number of Objaverse objects and randomly place them on top of the surface.
Objects are randomly scaled and placed in such a manner that they do not sit too close to the surface's margin.
We illuminate the scene evenly from the top, and add a overhead light source to cast shadows randomly.
Due to the highly stochastic nature of this process, it yields a large variety of samples, imposing very few restrictions on the content to reconstruct.

\paragraph{Processing \& export.}

We use an orthographic camera to render the scene in dimetric projection, mirroring the templates we generate (see \cref{sec:2d-template}).
We consider the unit cube, which surrounds the world origin, the \textit{core}.
To obtain the matching voxel grid, which serves as the training target, we crop a cube with a resolution of $M_\text{context} \times M_\text{context} \times M$ voxels around the origin (recall from \cref{sec:convolutional-sparse-structure-gen}).
This crop contains the core at its center (up to $M^3$ voxels) as well as additional voxels for context.
Following~\cite{xiang2024structured}, we render $\mathcal{V}_{\mathcal{F}}$ additional views using a perspective camera, extract DINOv2~\cite{oquab2023dinov2} features, and project them onto the voxel grid.
This yields a sparse structure and its corresponding structured latents.

\paragraph{Technical details.}

We implement the data engine using Blender.
We choose $\mathcal{V}_{\mathcal{F}} := 32 + 5$, where we generate $32$ random views of the scene from the top, a view from each side, as well as a view from the bottom.
\section{Experiments}%
\label{sec:experiments}

\paragraph{Experimental details.}

We use our dataset generation engine to produce 320k samples.
For fine-tuning the sparse structure and structured latent transformers of TRELLIS~\cite{xiang2024structured}, we use a learning rate of $5 \times 10^{-6}$, not freezing any layers, and a batch size of 1 (due to the additional voxels introduced).
The sparse structure model was fine-tuned for 260k steps, the structured latent model 660k steps on $2 \times$ NVIDIA RTX A6000 GPUs.2
The sparse structure model has a resolution of $(M_\text{context})^2 \times M = (M + 2V)^2 \times M = (16 + 2 \cdot 8)^2 \times 16$, the structured latent model $(N_\text{context})^2 \times N = (N + 2V)^2 \times N = (64 + 2 \cdot 32)^2 \times 64$.
The training objective task probability is set to $p = 0.5$.
The 2D inpainting model used during template generation is the Flux ControlNet of~\cite{alimama2024fluxcontrolnet}.
In the following, we present scenes that were obtained with layout instructions entirely generated by the large language model~(LLM) ChatGPT~5~Instant.
We choose a template window width of $896$~px (in the $2:1$~dimetric projection, the height is thus $448$~px), which corresponds to $M$ during the first stage and $N$ in the second stage. Put differently, this is the size of a single core patch in the template. Overall, the scene templates measure $1344 \times 672$~px.
The masks of the template windows are extruded by $60$~px. This imposes a maximum height of tall structures at the boundaries that we reconstruct. This value can be freely chosen but needs to be consistent in dataset generation, training, and scene inference.
During the convolutional inference, we use a stride of $0.5$ patches.

\paragraph{Comparisons.}

\begin{table}[t]
\centering
\caption{\textbf{Template faithfulness.} We evaluate multiple metrics to gauge the faithfulness of 3D scene reconstructions to the template with LPIPS~\cite{zhang18the-unreasonable}, structural similarity index measure (SSIM), and peak signal-to-noise ratio (PSNR). Our method with all proposed design choices leads to superior performance.}%
\resizebox{0.65\textwidth}{!}{%
\begin{tabular}{lccc}
\toprule
Method & LPIPS $\downarrow$ & SSIM $\uparrow$ & PSNR $\uparrow$ \\
\midrule
TRELLIS~\cite{xiang2024structured} & 0.4094 & 0.4966 & 13.5911 \\
TripoSG~\cite{li25triposg:} & 0.4182 & 0.4953 & 12.2768 \\
Hunyuan3D-2.1~\cite{hunyuan3d25hunyuan3d} & 0.4689 & 0.4490 & 11.5394 \\
\midrule
Ours (no fine-tuning) & 0.4726 & 0.4657 & 11.7622 \\
Ours (no context) & 0.4121 & 0.5142 & 14.2038 \\
Ours (smaller context) & 0.4000 & 0.5047 & 14.1266 \\
Ours (large stride) & 0.4143 & 0.5192 & 14.0624 \\
\midrule
Ours & \textbf{0.3993} & \textbf{0.5247} & \textbf{14.4137} \\
\bottomrule
\end{tabular}%
}%
\label{tab:perceptual}
\end{table}

\begin{figure}[t]
\includegraphics[width=\linewidth]{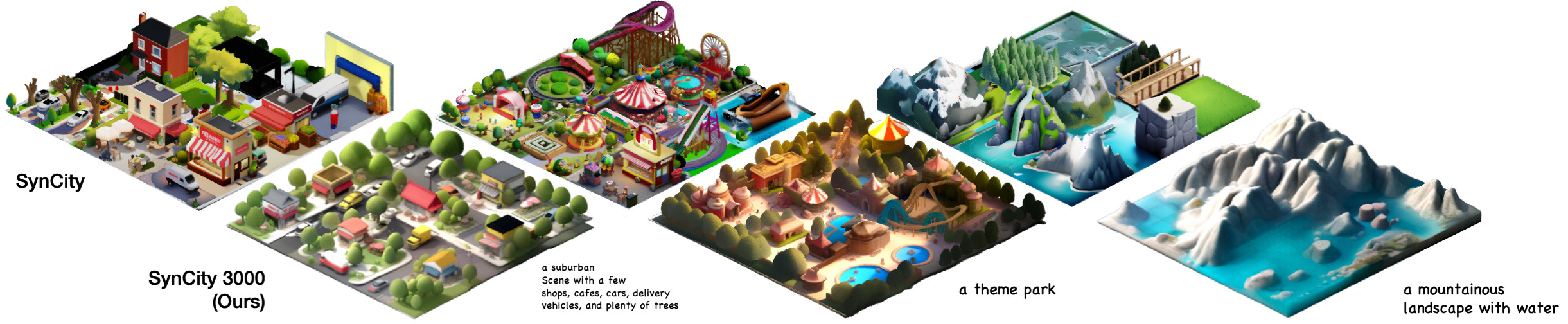}
\caption{\textbf{Comparison With SynCity~\cite{engstler2025syncity}.} We compare scenes generated by SynCity (left) and our method (right) using the same prompt for each. Overall, the worlds generated by our method are richer, more vibrant, and more coherent, both visually and semantically. They have an organic structure that lets elements flow across the whole scene.}%
\label{fig:syncity-comparison}
\end{figure}

We compare \method{} across multiple axes: layout control, faithfulness to the template, scene-scale reconstruction quality, and overall scene quality.
To this end, we conducted a user study (\Cref{tab:user-study}) and present multiple quantitative and qualitative results.

Unanimously, users prefer our flexible layout control over that of SynCity, which only allows rather rigid grids with a low resolution, severely limiting the creative freedom.
This becomes apparent in \cref{fig:syncity-comparison}, where we show qualitative comparisons to SynCity.
We query the same LLM to produce instructions with the same theme prompt.
SynCity creates scenes with a clearly apparent grid structure due to its tile-based generation mechanism.
While this approach works for scenes where this setting is natural (\eg, grid-planned cities), it falls short for those that are more open or have larger structures that span beyond a single tile (see the bottom row in \cref{fig:syncity-comparison}).
This is a methodological limitation of SynCity, which is fully alleviated by our convolutional approach.

\begin{figure}[t]
\includegraphics[width=\linewidth]{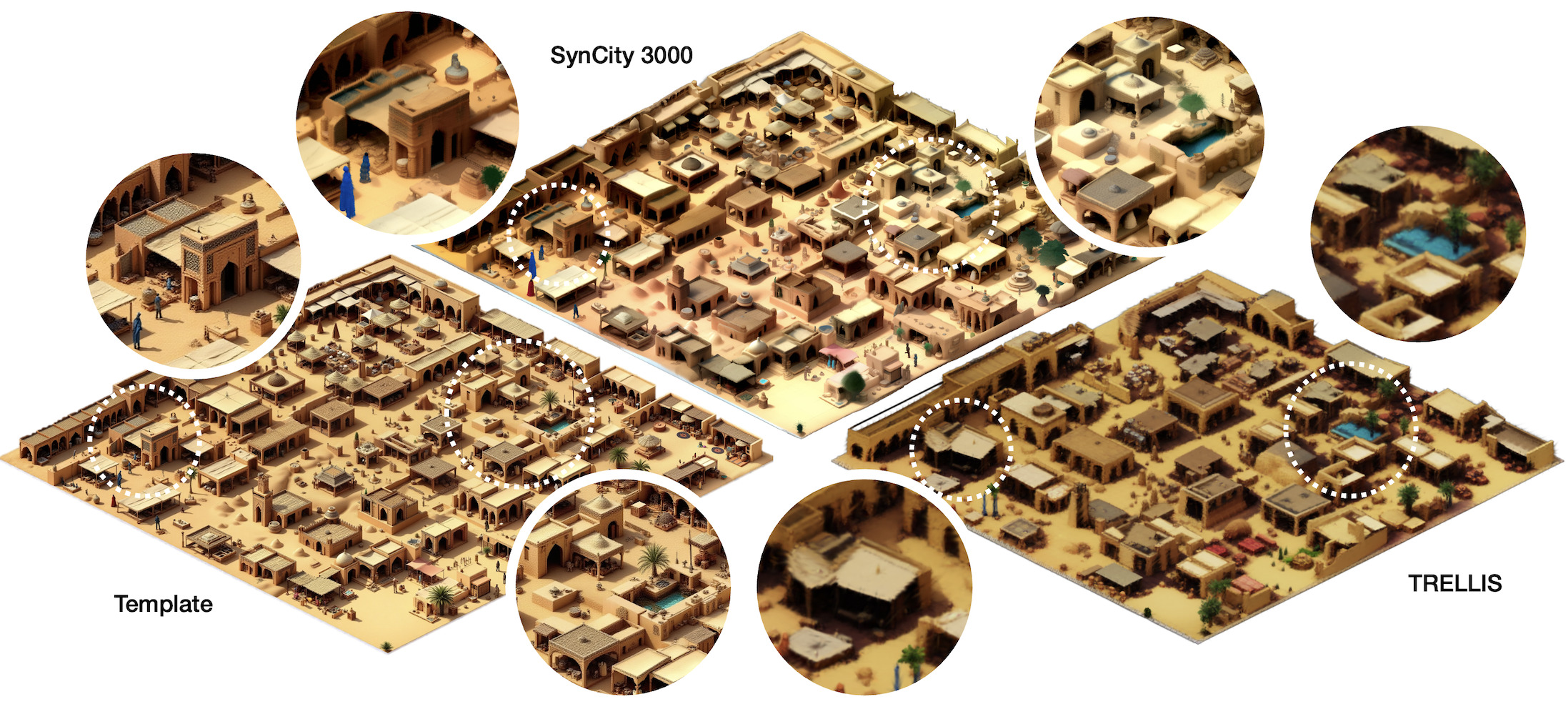}
\caption{\textbf{Large Scene Detail Comparison.}
We let both our method (top) and TRELLIS (right) reconstruct a larger scene ($3136 \times 1568$ pixels template, left).
While the overall scene produced by TRELLIS looks decent, its resolution notably impacts its capability to reproduce details.
Meanwhile, \method{} has a high adherence to the template and reproduces details more faithfully.
We recommend viewing on a display and zooming in.}%
\label{fig:recon-comparison}
\end{figure}

A key part of our method is accurately translating the generated 2D template into an actual 3D scene. Thus, we evaluate the faithfulness of the reconstructed scene to the templates next, comparing it with off-the-shelf image-conditioned models TRELLIS~\cite{xiang2024structured}, TripoSG~\cite{li25triposg:}, and Hunyuan3D-2.1~\cite{hunyuan3d25hunyuan3d}, as presented in \cref{tab:perceptual}.
We use a set of 35 templates, which were randomly generated using an LLM\@.
Based on this set, we generate scenes using all methods and render the results with an orthographic perspective that matches the template.
Then, we use image similarity metrics to evaluate the faithfulness of the reconstructions to their corresponding image templates.
We find that object-centric methods produce scenes at resolutions insufficient for this scale and thus lack finer details and occasionally deviate significantly from the conditioning image.
This becomes especially noticeable for larger scenes, as presented in \Cref{fig:recon-comparison}.
Please refer to the supplementary materials for additional qualitative comparisons.

The geometric quality of the reconstruction, apart from mere faithfulness to the template, also matters.
While scene generation is an inherently generative task with no ground-truth data, we can leverage our synthetic dataset engine to obtain scene-like proxies.
These allow us to gauge the geometric accuracy of the scene reconstructions as if ground truth were available.
To consider the effects of scene scale, we produce 25 scenes at two different scales, leading to different template sizes when rendered (one as used throughout this section, and one at $5 \div 3$ of that size).
For evaluating LPIPS and PSNR, we use 32 random views.
We report the results in \Cref{tab:geometric}.
While the performance of TRELLIS deteriorates as the scene size increases, ours stays consistent and ahead in most metrics.

For general generation preference, we asked users to pick a favorite scene in a binary forced-choice setting.
For fair comparison, we built scenes with prompts similar to those generated by NuiScene~\cite{lee25nuiscene:} and 3DTown~\cite{zheng25constructing}.
For SynCity, we used the same prompt (and applied no layout constraints), see \cref{fig:syncity-comparison}.
Scenes produced by \method{} were overall preferred.

While we evaluated template faithfulness and geometric quality, judging multi-view visual \textit{plausibility} is not as straightforward in the absence of ground-truth data.
Hence, we asked users to rate generated scenes from very implausible (1) to very plausible (5) and obtained an average rating of 3.57.

\begin{table}[t]
\centering
\caption{\textbf{User preference evaluation.} We asked users ($N = 27$) to state their preferred method in forced binary choices across different concerns: layout control, reconstruction faithfulness to the template, and their overall preferred scene. For each concern and method, we presented users with three scene renderings.}%
\resizebox{\textwidth}{!}{%
\begin{tabular}{c c c c c }
\toprule
\multicolumn{4}{c}{\textbf{Win Rate (\%)}}\\
\multicolumn{1}{c}{\textbf{Layout Control}} & \multicolumn{3}{c}{\textbf{Faithfulness of 3D Reconstruction to Template}}\\
SynCity~\cite{engstler2025syncity} & TRELLIS~\cite{xiang2024structured} & TripoSG~\cite{li25triposg:} & Hunyuan3D-2.1~\cite{hunyuan3d25hunyuan3d} \\
\midrule
100.0 & 71.6 & 100.0 & 100.0 \\
\midrule
\multicolumn{4}{c}{\textbf{Generated Scene Preference Win Rate (\%)}}\\
SynCity & NuiScene~\cite{lee25nuiscene:} & 3DTown~\cite{zheng25constructing} & TRELLIS w/ our template \\
\midrule
63.0 & 74.1 & 59.3 & 78.6 \\
\bottomrule
\end{tabular}
}%
\label{tab:user-study}
\end{table}

\begin{table}[t]
\centering
\caption{\textbf{Geometric reconstruction quality.} We evaluate the geometric reconstruction quality of TRELLIS~\cite{xiang2024structured} and our method on synthetic scenes using the Chamfer Distance, F-score, LPIPS, and PSNR\@.
We use the template size as described in our experimental details, as well as a larger one, with 25 scenes each.}%
\resizebox{0.65\textwidth}{!}{%
\begin{tabular}{lcccc}
\toprule
Method & Chamfer $\downarrow$ & F-score $\uparrow$ & LPIPS $\downarrow$ & PSNR $\uparrow$ \\
\midrule
\multicolumn{5}{c}{Template Size: $1344 \times 672$~px} \\
\midrule
TRELLIS & \textbf{0.0137} & 0.6685 & 0.5628 & 12.2615 \\
Ours & 0.0166 & \textbf{0.7029} & \textbf{0.5340} & \textbf{13.3197} \\
\midrule
\multicolumn{5}{c}{Template Size: $2240 \times 1120$~px} \\
\midrule
TRELLIS & 0.0392 & 0.6967 & \textbf{0.5762} & 11.6361 \\
Ours & \textbf{0.0302} & \textbf{0.7536} & 0.5812 & \textbf{12.6063} \\
\bottomrule
\end{tabular}%
}%
\label{tab:geometric}
\end{table}

\paragraph{Ablations.}

We further show the effectiveness of our design choices in \cref{tab:perceptual}.
Central to our method is the fine-tuning of models presented in~\cite{xiang2024structured} to enable convolutional inference.
By using the models off the shelf, performance suffers significantly.
This can be mainly attributed to the fact that the sparse structure model tries to infer a canonical front-facing orientation, which is ill-defined for objects that lack this directionality.
Without context, there is a minor deterioration in quality, indicating that the model benefits from information about the surrounding voxels.
This is also true if we reduce the context ($V=4/16$ for stage 1 and 2, respectively), indicating that the model captures long-range dependencies.
If we use a large stride (here, an entire patch, which leads to no averaging of predictions), the reconstructions are also inferior.
Smaller steps, and thus averaging tile predictions, help resolve ambiguities arising from template masking.
For a single patch, structures in the background may be visible if they are not occluded by those in the foreground.
This can lead to duplicated structures, whose occurrence is significantly reduced with smaller strides.

\section{Conclusion}%
\label{sec:conclusion}

We have introduced \method, a method to generate large-scale 3D scenes from scratch.
Starting from a straightforward approach to describe arbitrarily complex scene layouts, we use a shifted-window approach to turn them into 2D templates, which are then converted into 3D Gaussian Splats through convolutional inference by a fine-tuned two-stage 3D generative model.
We fine-tune this model with data generated by a synthetic dataset engine, which creates random scene-like 3D structures.
By first generating the complete scene as a template in a joint diffusion process, we ensure semantic and visual coherence, directly addressing the limitations of SynCity.
Furthermore, by fine-tuning the 3D generative model, we maintain its high-quality predictions while eliminating the need for brittle heuristics required by previous methods for scene generation.

\paragraph*{Ethics.}

For details on ethics, data protection, and copyright, please see \url{https://www.robots.ox.ac.uk/~vedaldi/research/union/ethics.html}.

\paragraph*{Acknowledgments.}

The authors of this work are supported by ERC 101001212-UNION and Meta Research.
Special thanks to Robin Y. Park for her technical contributions.

\bibliographystyle{splncs04}
\bibliography{vedaldi_general,vedaldi_specific,main}
\clearpage
\setcounter{page}{1}
\appendix

\title{SynCity 3000: Bootstrapping Scene-Scale 3D Diffusion (Supplementary Materials)}
\titlerunning{SynCity 3000}

\author{
Paul Engstler
\and
Iro Laina
\and
Christian Rupprecht
\and
Andrea Vedaldi}
\institute{
Visual Geometry Group, University of Oxford \\
\email{\tt\small \{paule,iro,chrisr,vedaldi\}@robots.ox.ac.uk}
}
\authorrunning{P.~Engstler et al.}

\maketitle

\section{Additional technical details}

\paragraph{LLM prompting.}

While the layout instructions can be created manually, high-quality results can be easily obtained by using an LLM\@.
This is the prompt we have used for ChatGPT~5~Instant:

\begin{quotation}
\noindent
Imagine you are able to generate small worlds described by JSON files.
The scales of these worlds could range from a small town to a region in a park.
I need you to define an overall scene prompt, which specifies the general concept of the scene in very easy words, then a theme prompt, which gives a stylistic hint (that should work for diffusion models).
We can also define constraints, which can be used to describe specific regions in more detail (the positions are specified with x and y coordinates and width and height).
For example, objects may be ``forced'' to appear there.
I'd like you to get creative now and design a few more worlds.
Use at most three constraints (preferably fewer, or even none).
Do not change the fields \texttt{scene\_size}, \texttt{window\_size}, \texttt{stride}, or \texttt{extrusion\_height}.
Avoid the mention of fog, steam, clouds, and (very) small objects.
Here is an example: \texttt{<EXAMPLE JSON>}.
Now, create a few additional worlds of various styles.
\end{quotation}

Note that to simplify the instructions for the model, we use relative patch units here as opposed to pixels (see the implementation details in \cref{sec:experiments}).
Furthermore, these instructions are only used for the 2D template generation.

\begin{figure}[ht]
\begin{lstlisting}[language=json,basicstyle=\small\ttfamily,breaklines=true,columns=fullflexible,tabsize=2]
{
  "scene_prompt": "a theme park",
  "scene_size": 3,
  "window_size": 2,
  "stride": 1,
  "theme_prompt": "uniform noon lighting, 8k, realistic textures, subtle gradients, isometric perspective",
  "constraints": [
    {
      "x": 0.3,
      "y": 0.5,
      "width": 1.5,
      "height": 1,
      "prompt": "a wooden rollercoaster"
    },
    {
      "x": 1,
      "y": 2.25,
      "width": 0.5,
      "height": 0.5,
      "prompt": "a merry-go-round"
    }
  ],
  "extrusion_height": 60
}
\end{lstlisting}
\vspace{-1em}
\caption{Example JSON file to describe a theme park scene.}%
\label{fig:lang-prompt}
\end{figure}

\paragraph{Latent encoding \& decoding.}

During dataset generation, we encode voxel data, and during inference with our pipeline, we decode the structured latents into 3DGS\@.
For these tasks, we use the variational autoencoders~(VAE) presented in~\cite{xiang2024structured} off the shelf.
As mentioned in Section~3.2.1, the sparse structure VAE is fully convolutional.
Thus, we can already use it for scenes of any size.
However, the structured latent VAE and the 3DGS decoder head are transformer-based.
For them, we employ a shifted-window approach to make them work for scenes of any size.
When decoding the structured latents, we found that using a shifted-window approach with a small stride and averaging yields the best results.
For decoding the 3DGS, averaging them is not straightforward.
Thus, we predict Gaussians for each voxel exactly once and fuse them together to create a large scene representation.

\paragraph{Color correction.}

Reconstructions generated by TRELLIS~\cite{xiang2024structured} (without any of our proposed fine-tuning) show a reduction in lightness.
This can be observed in~\cref{fig:recon-comparison}, where its reconstructions appear to not be as brightly lit as the template.
We conjecture that this issue arises either from the way that the training data is being generated or from an issue with the latent code learned by the structured latent generation model and the subsequent decoding.
During training data generation, random views of an object are encoded with DINOv2~\cite{oquab2023dinov2}.
After voxelizing the object, these features are then projected onto the voxels and accumulated.
Crucially, voxel visibility is not taken into account.
Thus, every voxel along a ray receives that pixel's feature.
By averaging the feature contributions from a large number of random views, the assumption is that voxels will eventually approach their \emph{true} features.
However, especially for complex objects with little visual symmetry, this might not be true.
As we merely fine-tune the structured latent model and use the original decoders, we are bound to follow this same setup (see~\cref{fig:dataset}).

To compensate for this shortcoming, we transfer the color statistics (mean and standard deviation for each channel) from the template to the Gaussians in the L*a*b* color space~\cite{ISO11664}, which considers perceptual luminance.
For the colors of the Gaussians to look \emph{correct}, our method has to still predict colors correctly in its less illuminated color space, as we only transfer the statistics.
Thus, this is merely a visual enhancement and by no means a simplification.

\paragraph{Template generation details.}
For clarity, we have applied minor simplifications to the template generation process outlined in Section 3.1, which we explain here in more detail.
First, the extent of $I_{base}$ and $M_{base}$ depends on the size of the window (or layout constraint) being used, as they directly describe part of the generated image (or corresponding latent) and thus need to match its size.
Second, to correctly average the noise over all windows, we determine the number of overlapping windows for each latent \textit{pixel} and weight the windows uniformly.
Third, if we only masked windows, the non-window background of latents would not be conditioned at all and would appear noisy.
To remedy this, we ensure that (non-layout-constraint) windows also contribute to the (non-relevant) template background that they cover.
As implied in~\cref{fig:2d-gen}, we use a black background.
For visualization purposes, we remove this background throughout the paper.

\paragraph{Inference.}
When generating the template for a scene with Flux and the ControlNet of~\cite{alimama2024fluxcontrolnet}, we use a classifier-free guidance scale of 3.5 and 24 inference steps per window.
The vast majority of the scenes presented in figures took about 3--5 minutes to generate with a single NVIDIA RTX A6000, depending on the number of constraints.

We use slightly different configurations for the two stages in the 3D generation process.
For the sparse structure, the guidance scale is set to 7.5 and we sample 50 steps.
This choice slows down the inference but leads to improved results.
However, if minor qualitative degradations are acceptable, this number can be dropped significantly.
For the structured latents, we use a guidance scale of 3.0 and 12 sampling steps.
Taken together, these two stages take about 30 minutes to infer an average scene we present in this paper with a single NVIDIA RTX A6000.

These numbers are reported for the settings given in the \textit{Experimental Details} section.
For larger templates and scenes (such as those presented in \cref{fig:supp-larger-scene}), these numbers scale roughly quadratically, as expected in a convolutional operation.
The runtime and peak memory at template sizes (px) $1344 \times 672$, $2240 \times 1120$, and $3136 \times 1568$ on a single NVIDIA H100 GPU are 31m/45.7G, 82m/48.4G, and 179m/55.8G.

For creative processes with a human in the loop, the speed of the template generation is significantly more important.
Our 3D generation process is purely transforming the template into Gaussian splats without any further modifications (and with smaller quality degradation than prior methods, as shown in \cref{tab:perceptual}).

\section{Additional experimental details}
To compare reconstruction faithfulness to a given template across methods, we used 35 templates whose prompts were randomly generated with an LLM\@.
These span six themes (fortress, mountains, solarpunk city, suburban town, theme park, and university campus).
We used seven different seeds for each theme and then ran our methods on all these templates (42 templates in total).
As TripoSG and Hunyuan3D-2.1 failed for 7 scenes, we removed them from the evaluation and compared only the remaining 35 templates.
Please see~\cref{fig:supp-full-comparison} for examples of the templates we used.
We presented them in this way so users could pick their preferred method for reconstruction faithfulness.

To evaluate the overall scene preference in our user study, we built three scenes each to compare prior works with ours.
Here, we paid special attention to the abilities of each method as well as the availability of code and data to enable fair comparisons.
For SynCity, we used the scenes shown in~\cref{fig:syncity-comparison}, which balance scenes with grid-like as well as non-grid-like structures, avoiding selection bias toward either method.
As NuiScene does not support user input but instead relies on its training data to produce unbounded scenes of a certain style, we picked prompts for \method{} that produce similar scenes (fortress, suburban city, and pagodas).
We further present these scenes with textures, even if this texturing is not directly part of NuiScene.
Due to the lack of published code, we took the three scenes in the teaser figure of 3DTown and converted them into a template suitable for our method with Nano Banana Pro.

\section{Limitations}
While our method shows significant improvement over previous works, there are still some limitations to be addressed in future work.

\paragraph{Forced perspective.}
While our method is not limited by the scene it can reconstruct, the convolutional inference setup requires the scene to be presented in a dimetric perspective in the template.
Future work could consider finding an approach that does not force a certain perspective.

\paragraph{Texture detail.}
As we use overlapping shifted windows, we average the results of the structured latent model.
Therefore, the fidelity of the textures is slightly reduced compared to the original TRELLIS model.

\paragraph{Degree of control.}
While the method we propose to generate templates provides a high degree of control, users still need to wait until the template has been produced to see if they are satisfied with the result.
Future work could investigate more interactive or even real-time approaches that allow users to intervene directly.

\paragraph{Structure duplication.}
As part of the convolutional design of our method, we tile the template.
This requires our fine-tuned stage 1 model to correctly align the geometry it is reconstructing.
Occasionally, with tall or large structures and in the presence of severe occlusion, this can lead to duplicated or partially broken structures.

\paragraph{Cartoonish appearance.}
Owing to biases in the training of TRELLIS, which we suspect could stem from the Objaverse-XL dataset, which includes stylized (and thus not necessarily realistic) objects designed by humans, and FLUX, which likely infuses additional biases due to the dimetric nature of scenes, usually reminiscent of video games, the scenes exhibit a cartoonish look.
This is also observed in SynCity.
By using Objaverse-XL for fine-tuning (and maintaining the training setup of TRELLIS), we presume these biases to be amplified.
Thus, our method has limited ability to replicate real-world scenes realistically.

\section{Additional comparisons}

To augment the quantitative results we present in Table 1, where we show perceptual metrics for the 3D reconstruction capability of off-the-shelf image-conditioned models TRELLIS~\cite{xiang2024structured}, TripoSG~\cite{li2025triposg}, and Hunyuan3D-2.1~\cite{hunyuan3d2025hunyuan3d}, we provide qualitative comparisons in \cref{fig:supp-full-comparison}.

TripoSG and Hunyuan3D-2.1 struggle to reconstruct a scene that resembles the template.
Aside from poor color accuracy, the geometry itself is not reliably reconstructed and may exhibit holes.

Overall, the reconstructions from TRELLIS look significantly more convincing, but compared to our method, there are still notable differences.
First, TRELLIS has a limited output resolution where details appear fuzzy or blurry.
Our method is scalable both in terms of scene size and resolution.
Second, TRELLIS generally follows the template quite well, but there are severe deviations at times, especially when considering details.
Our method reliably reconstructs the overwhelming majority of details (see the bottom row, where our method maintains car positions, orientation, and shape).

Despite the strong performance of recent image-conditioned 3D generative models, our fine-tuning approach yields a model that achieves more faithful reconstructions down to the details.
Further, our method can scale to arbitrary scene sizes, whereas the other models are limited in output size due to their object-centric design.

\section{Additional qualitative results}

In the following, we present further qualitative results of \method.
We would also like to invite the reviewers to consider the videos of additional scenes in the supplementary material.

\paragraph{Scene size.} One of the parameters that can be set in the layout instructions is the scene size.
This setting influences the template size and, by extension, the size of the 3D scene.
While \method{} can output scenes of arbitrary complexity even when the scene size remains fixed, this parameter increases the ``resolution'' of the scene, improving detail fidelity.
The scenes in \cref{fig:supp-larger-scene} have a template size of $3136 \times 1568$ pixels.
Consider the loss of details in scenes generated by TRELLIS due to its lower resolution.

\begin{figure}[t]
\centering
\includegraphics[width=\linewidth]{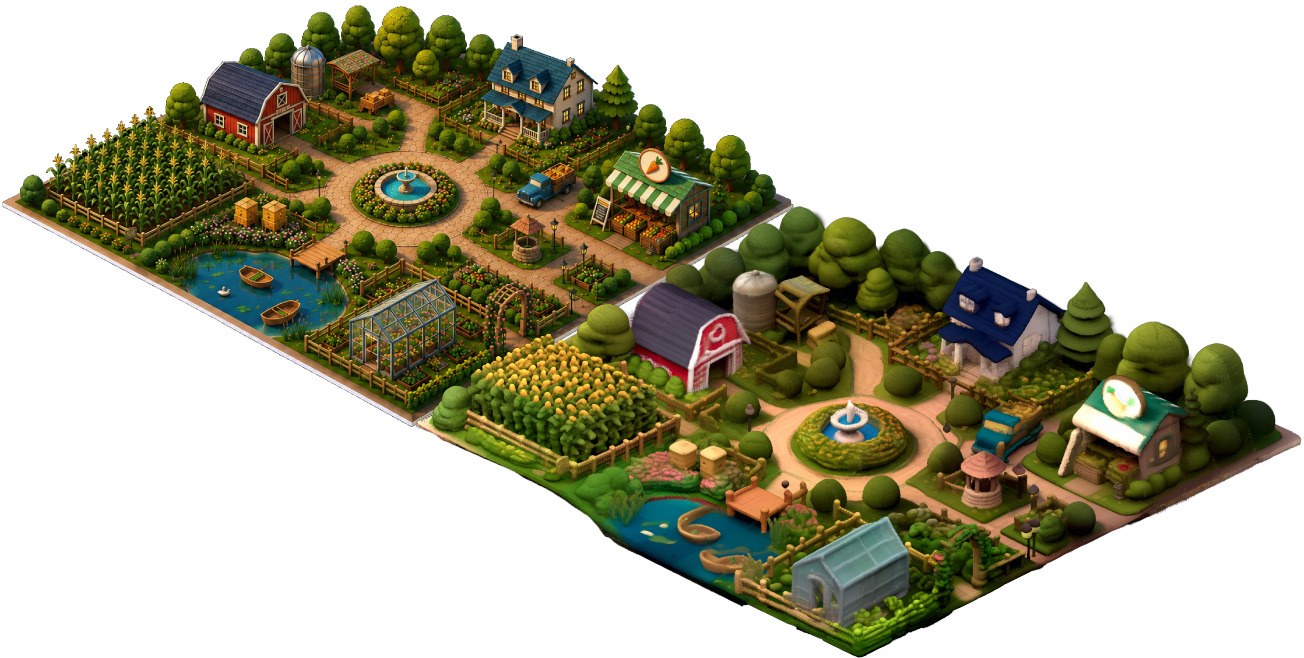}
\caption{A scene template generated by ChatGPT Images (left) and the resulting scene produced by our method (right).}%
\label{fig:gpt-template}
\end{figure}

\paragraph{Templates not generated by our pipeline.}
Our synthetic dataset engine creates highly diverse samples that are not tied to the results of our scene template generation pipeline.
Thus, we can produce scenes with our fine-tuned version of TRELLIS that were not generated with our pipeline, as shown in \cref{fig:gpt-template}.

\begin{figure}[t]
\centering
\includegraphics[width=\linewidth]{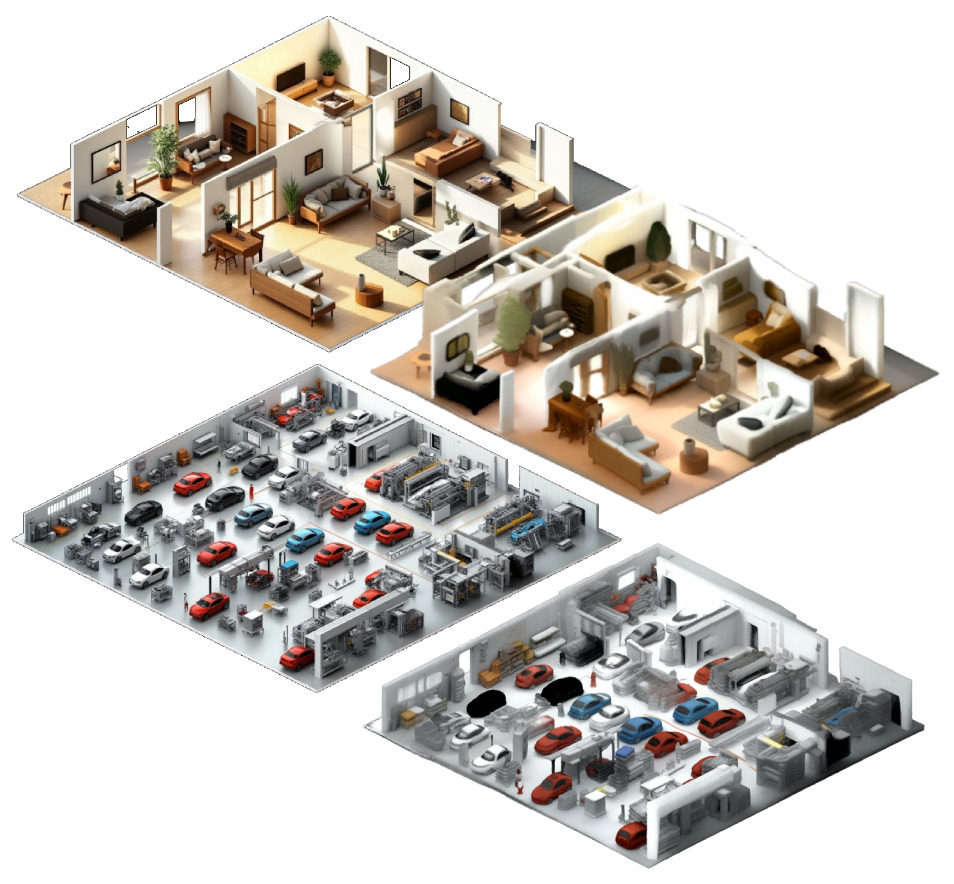}
\caption{Indoor scenes generated by our method (left template, right reconstruction).}%
\label{fig:supp-indoor-scenes}
\end{figure}

\begin{figure*}[t]
\centering
\includegraphics[width=\linewidth]{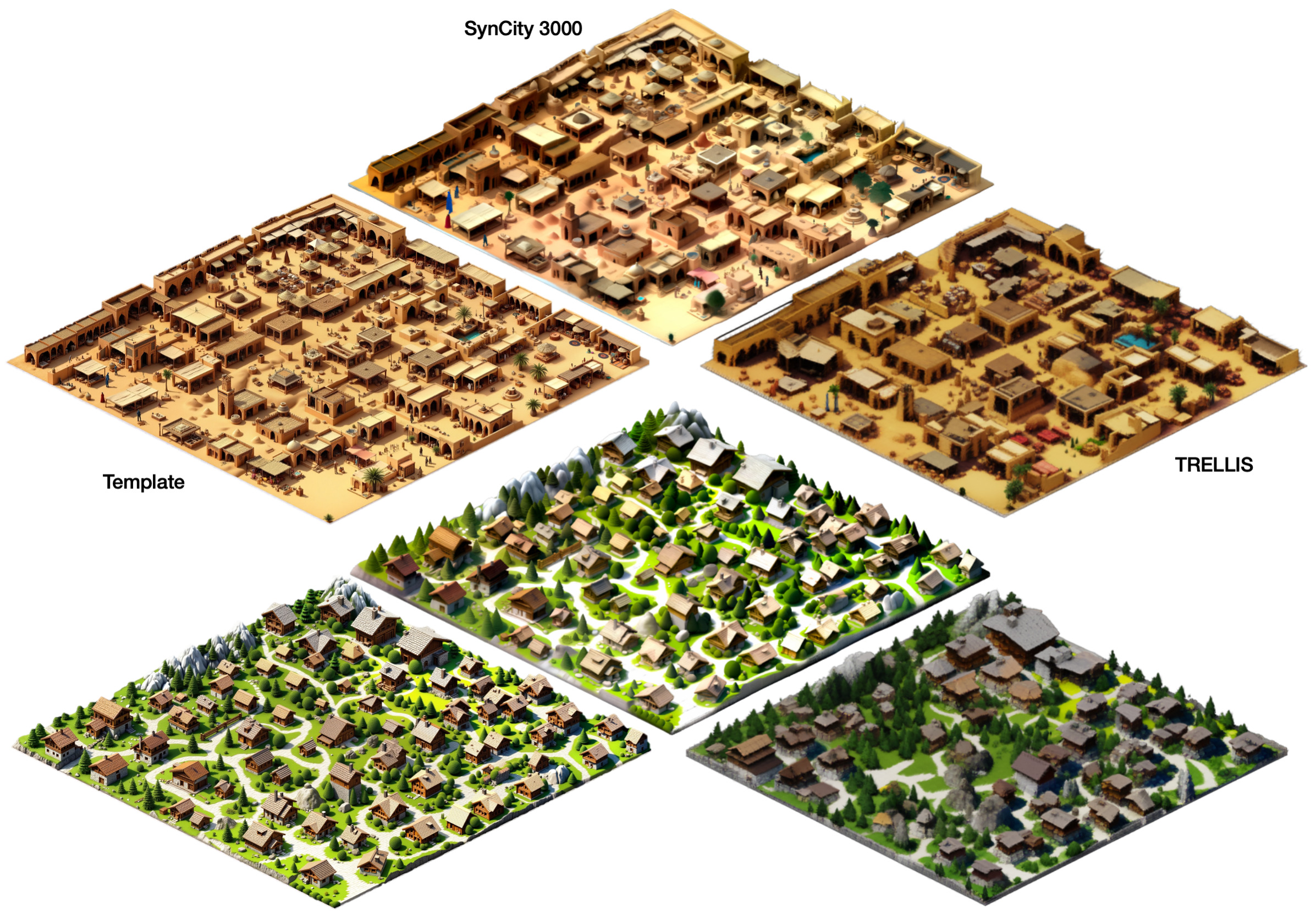}
\caption{Two large scenes, shown as templates (left), as generated by our method (top) and TRELLIS (right).}%
\label{fig:supp-larger-scene}
\end{figure*}

\begin{figure*}[t]
\centering
\includegraphics[width=\linewidth]{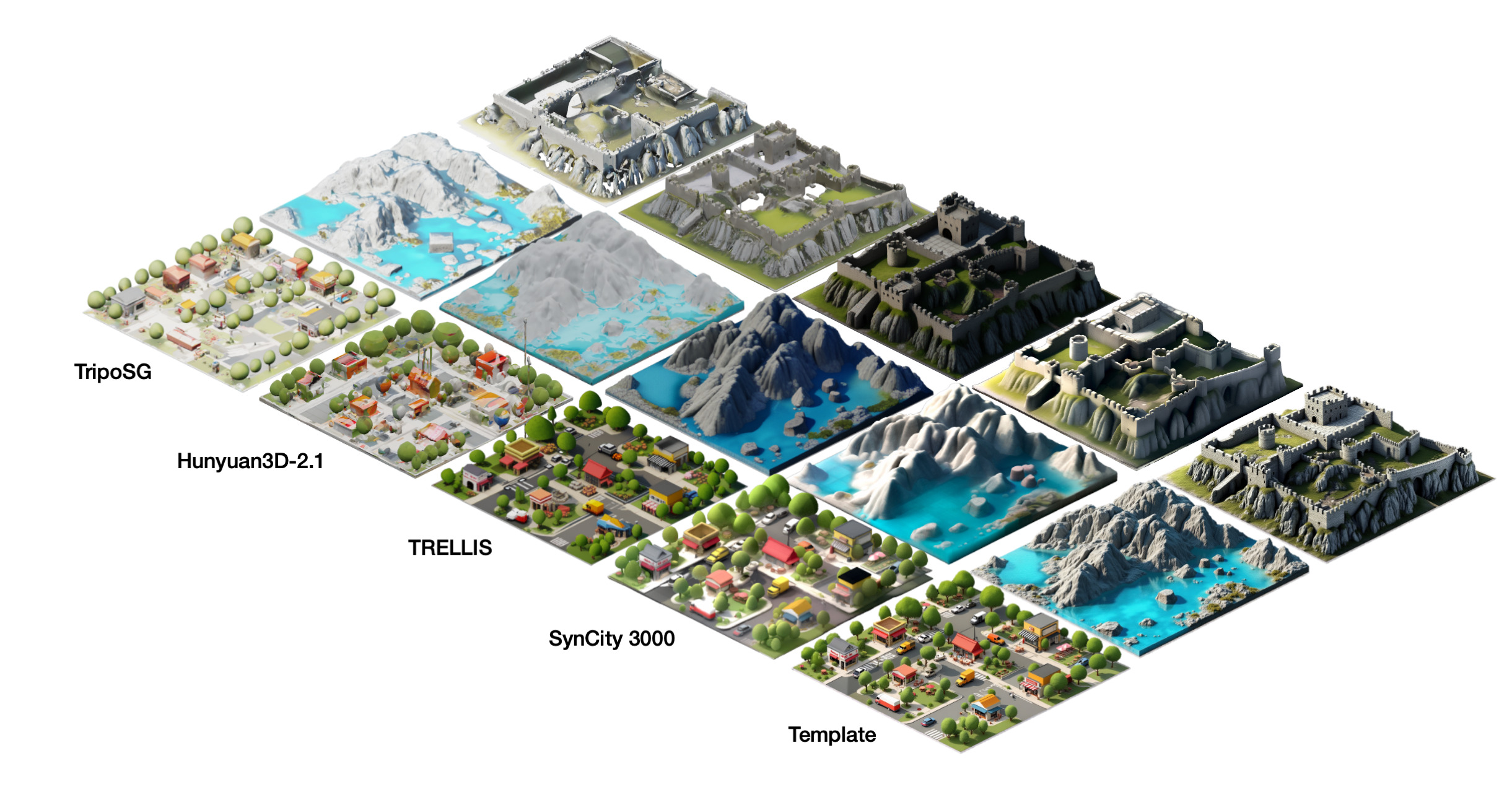}
\caption{Qualitative examples comparing TripoSG~\cite{li2025triposg}, Hunyuan3D-2.1~\cite{hunyuan3d2025hunyuan3d}, TRELLIS~\cite{xiang2024structured}, and \method.}%
\label{fig:supp-full-comparison}
\end{figure*}

\paragraph{Indoor scenes.}
While the prior work we compare to generates open outdoor scenes, which we replicate to ensure a fair comparison, our fine-tuning setup is technically agnostic to scene design.
However, we rely on FLUX to build plausible scene templates.
Due to its biases, the quality and plausibility of indoor scenes are limited, as shown in \cref{fig:supp-indoor-scenes}.
Improving both aspects is an interesting avenue for future work, as indoor scene generation is another highly active research domain~\cite{tang2024diffuscene, meng2025lt3sd, zhai2024echoscene, lin2024instructscene}.

\begin{figure*}[ht]
\centering
\includegraphics[width=\linewidth]{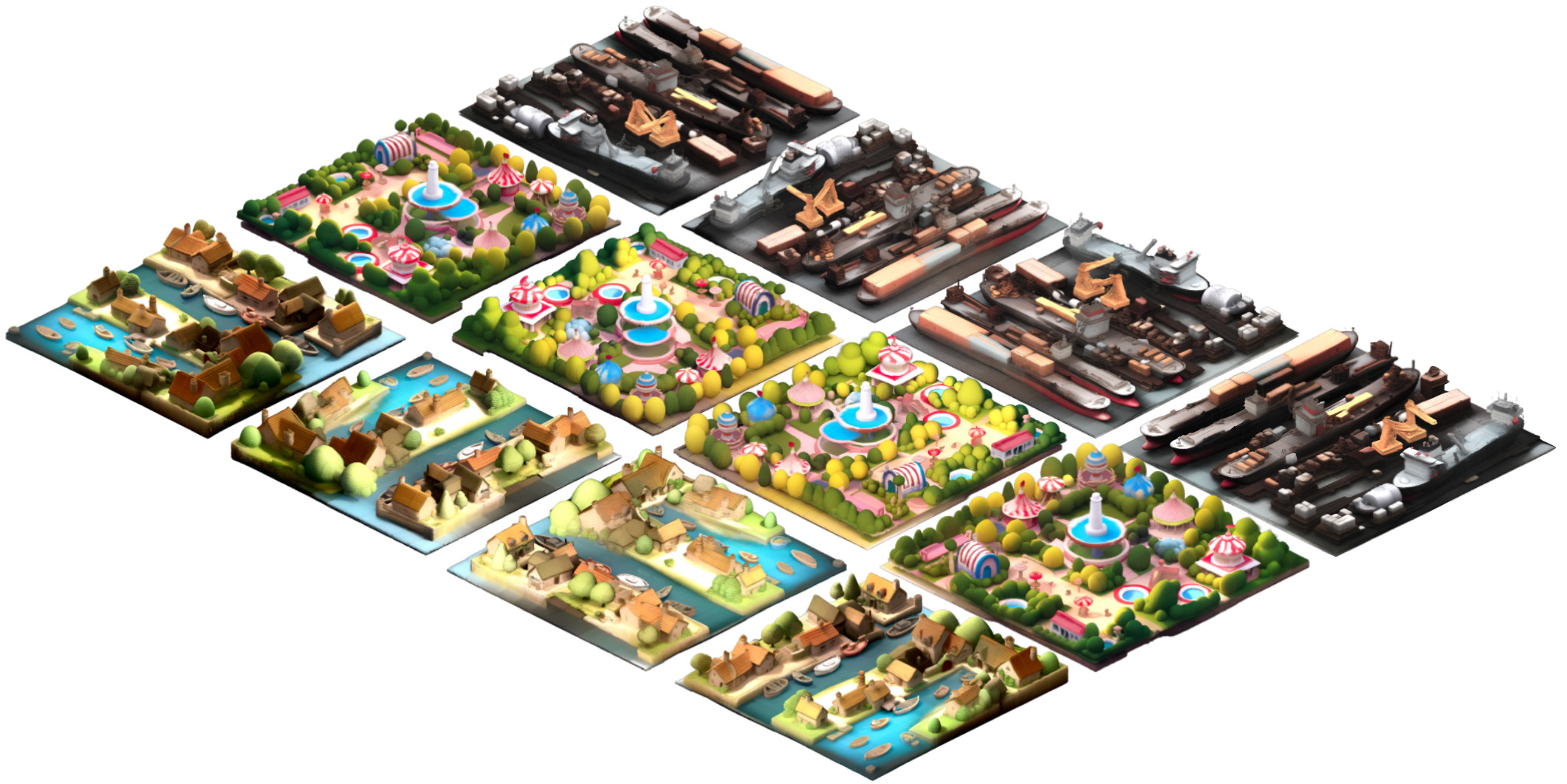}
\caption{Three scenes generated by \method, shown from all viewpoints.}%
\label{fig:supp-all-views}
\end{figure*}

\paragraph{Additional views.}
In \cref{fig:supp-all-views}, we present additional views of generated 3D scenes to show them in their completeness.
The scenes generated by \method{} are fully realized and do not exhibit holes compared to prior work~\cite{zheng2025constructing}.

\begin{figure*}[ht]
\centering
\begin{tabular}{ccccc}
\includegraphics[width=0.180\textwidth]{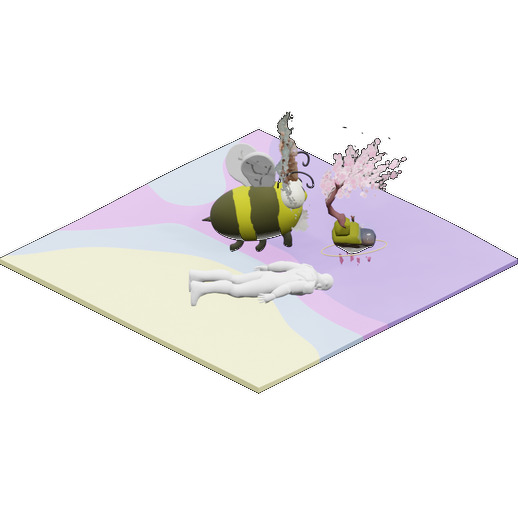} & \includegraphics[width=0.180\textwidth]{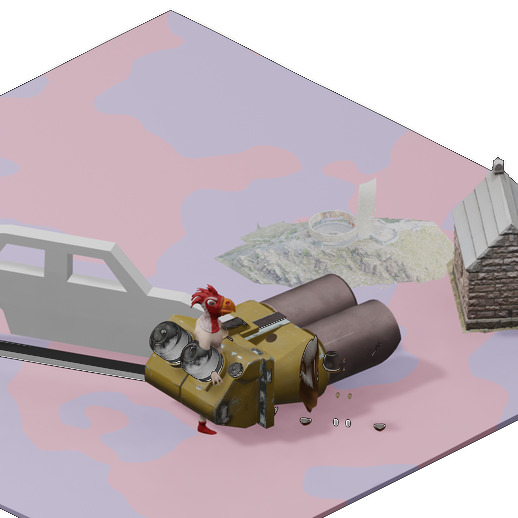} & \includegraphics[width=0.180\textwidth]{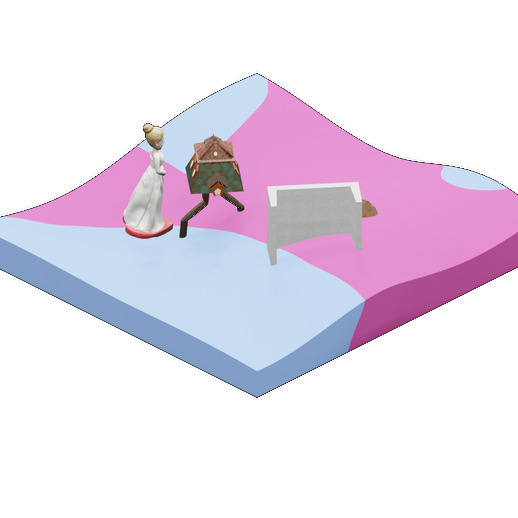} & \includegraphics[width=0.180\textwidth]{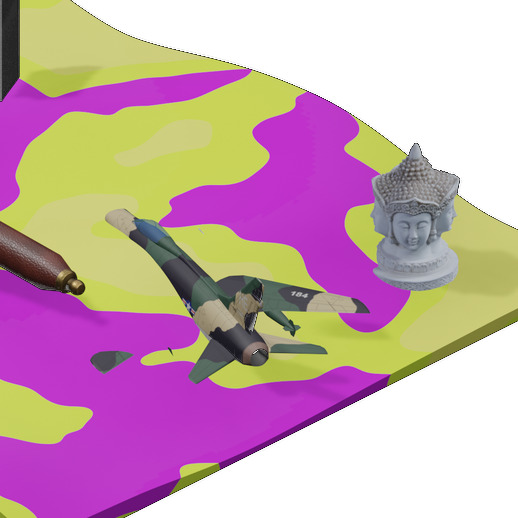} & \includegraphics[width=0.180\textwidth]{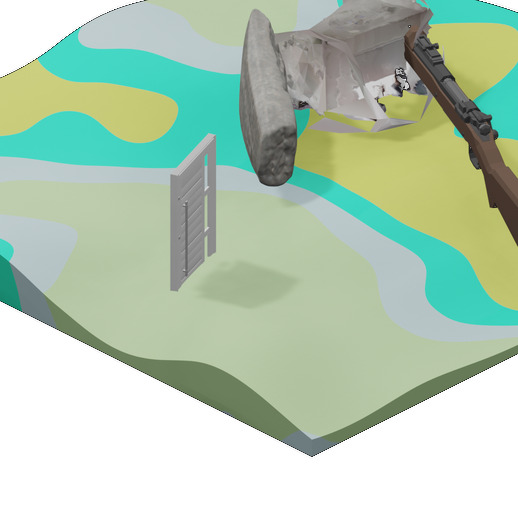} \\

\includegraphics[width=0.180\textwidth]{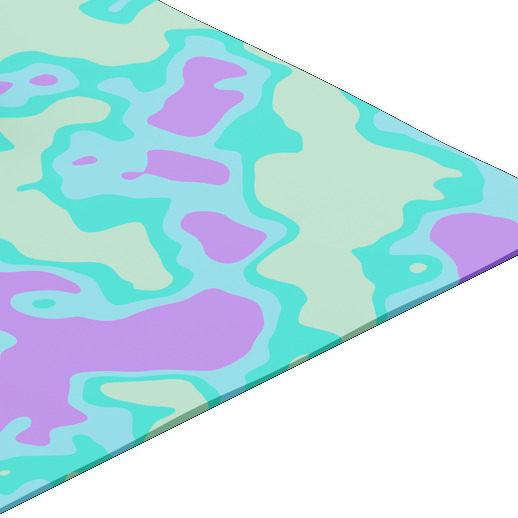} & \includegraphics[width=0.180\textwidth]{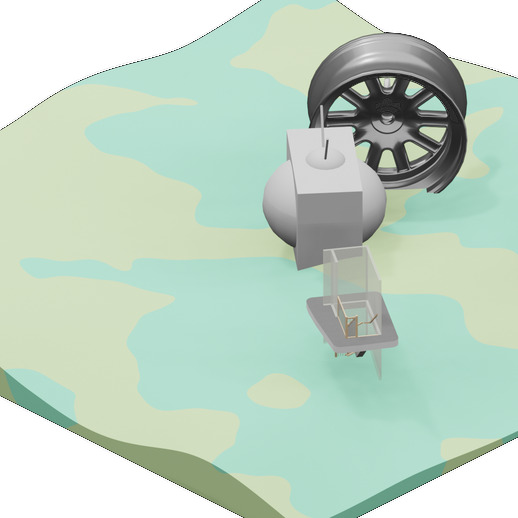} & \includegraphics[width=0.180\textwidth]{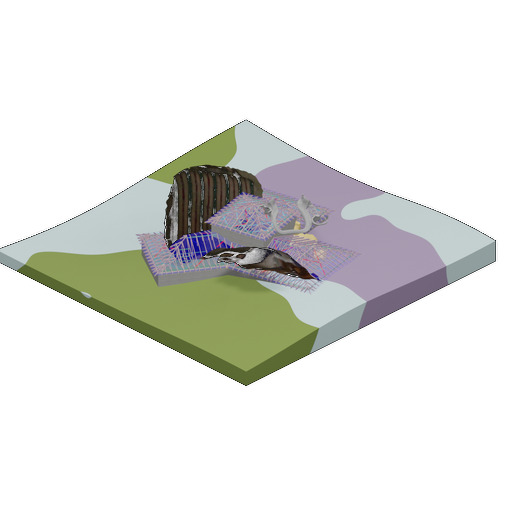} & \includegraphics[width=0.180\textwidth]{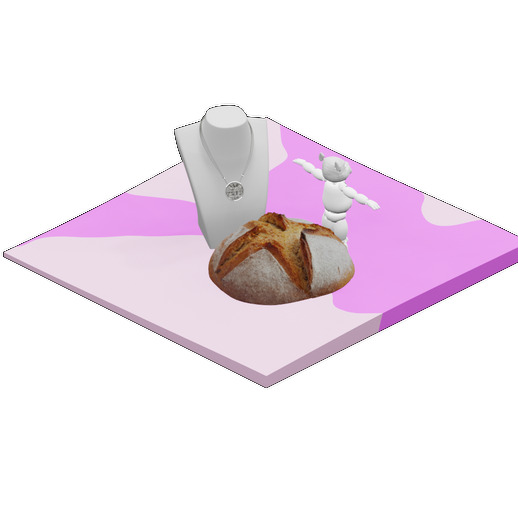} & \includegraphics[width=0.180\textwidth]{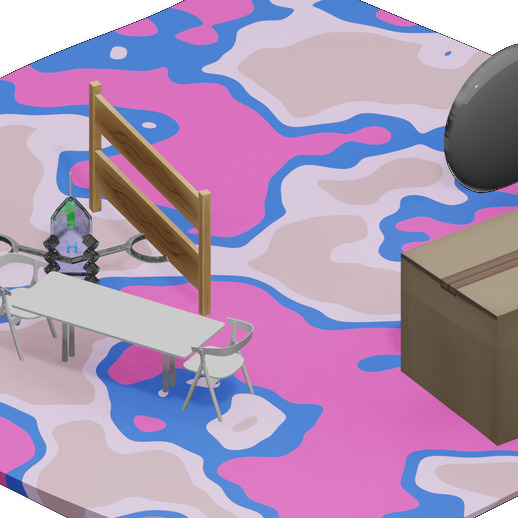} \\

\includegraphics[width=0.180\textwidth]{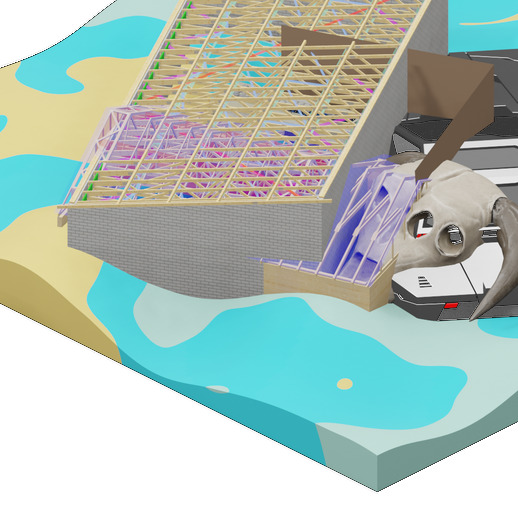} & \includegraphics[width=0.180\textwidth]{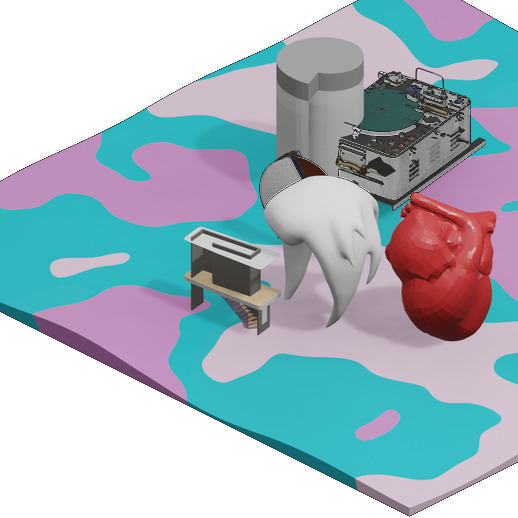} & \includegraphics[width=0.180\textwidth]{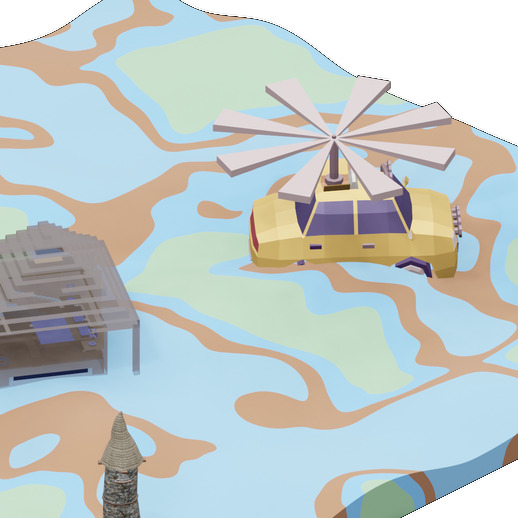} & \includegraphics[width=0.180\textwidth]{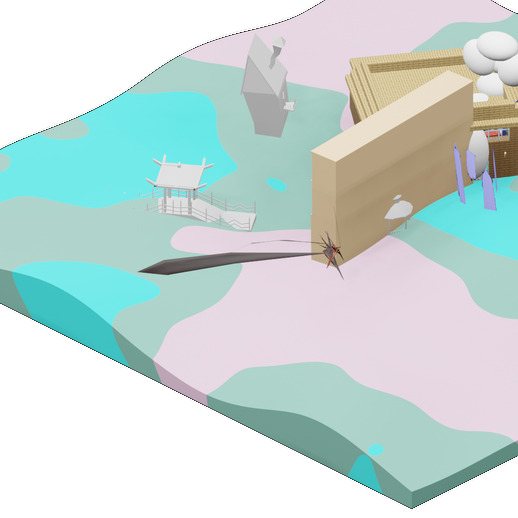} & \includegraphics[width=0.180\textwidth]{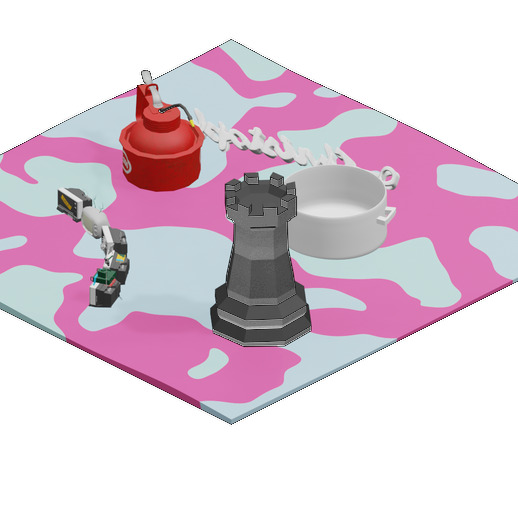} \\

\includegraphics[width=0.180\textwidth]{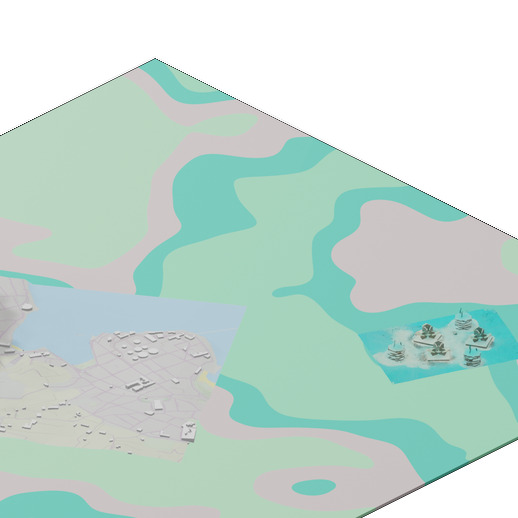} & \includegraphics[width=0.180\textwidth]{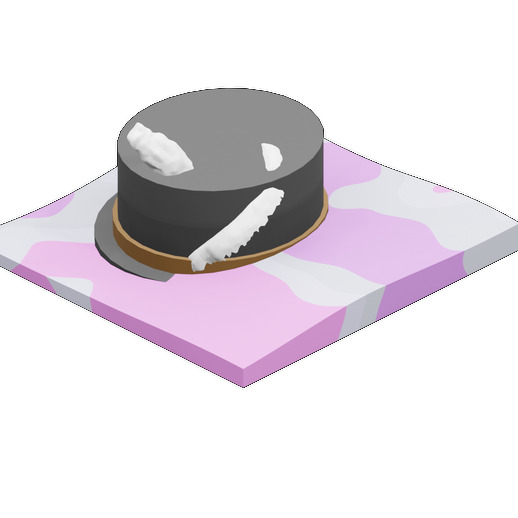} & \includegraphics[width=0.180\textwidth]{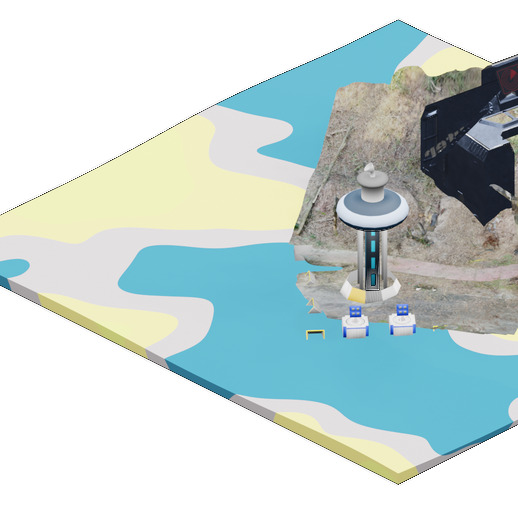} & \includegraphics[width=0.180\textwidth]{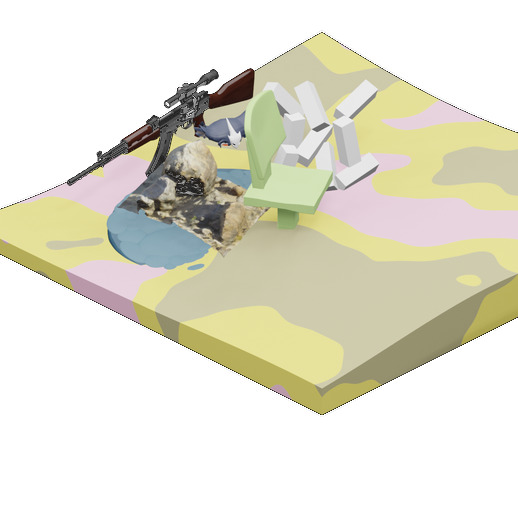} & \includegraphics[width=0.180\textwidth]{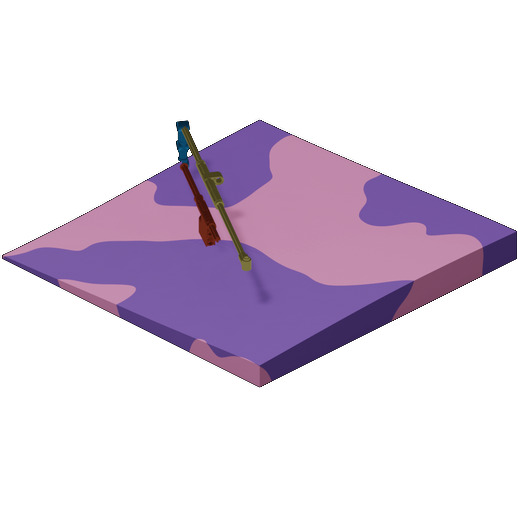} \\

\includegraphics[width=0.180\textwidth]{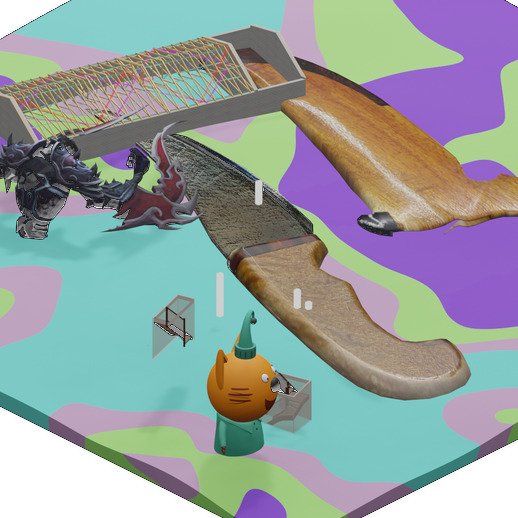} & \includegraphics[width=0.180\textwidth]{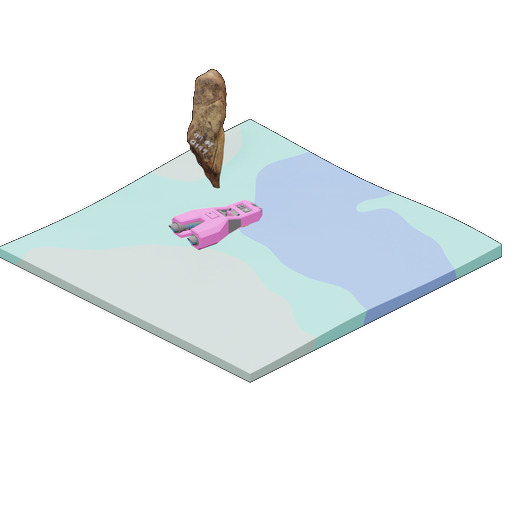} & \includegraphics[width=0.180\textwidth]{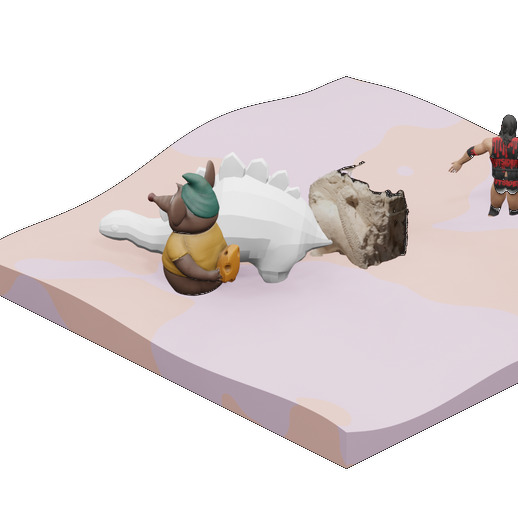} & \includegraphics[width=0.180\textwidth]{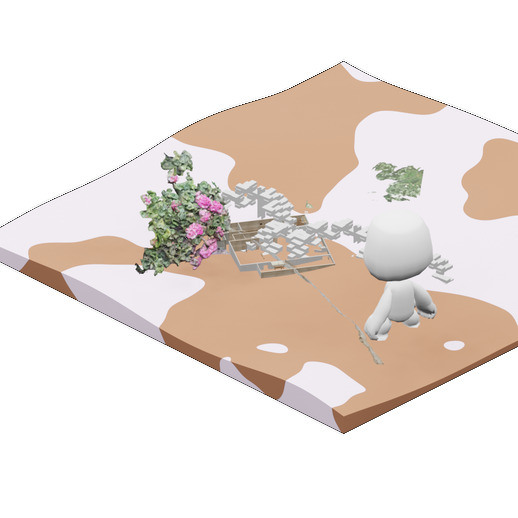} & \includegraphics[width=0.180\textwidth]{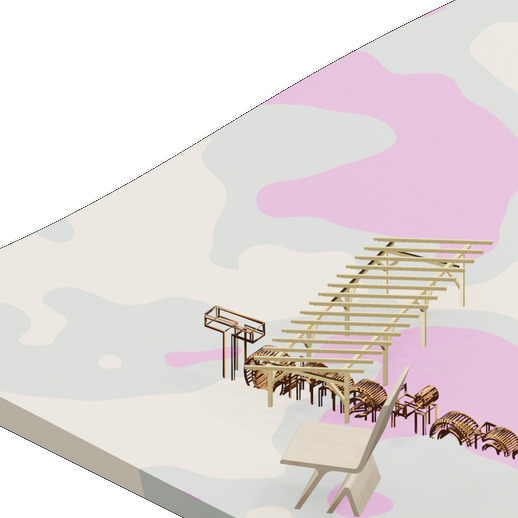} \\

\includegraphics[width=0.180\textwidth]{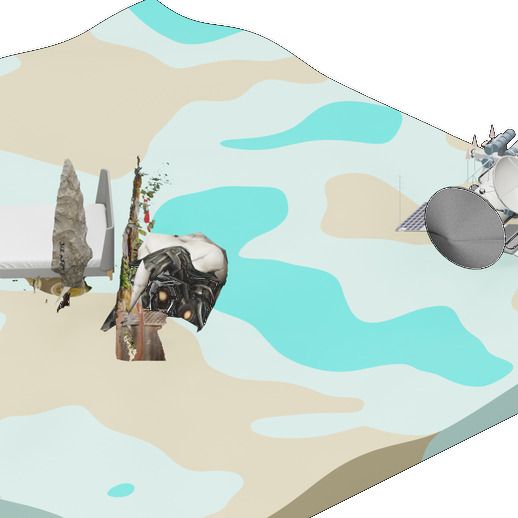} & \includegraphics[width=0.180\textwidth]{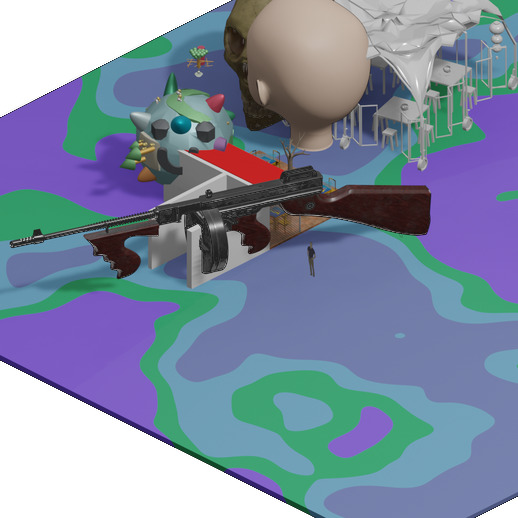} & \includegraphics[width=0.180\textwidth]{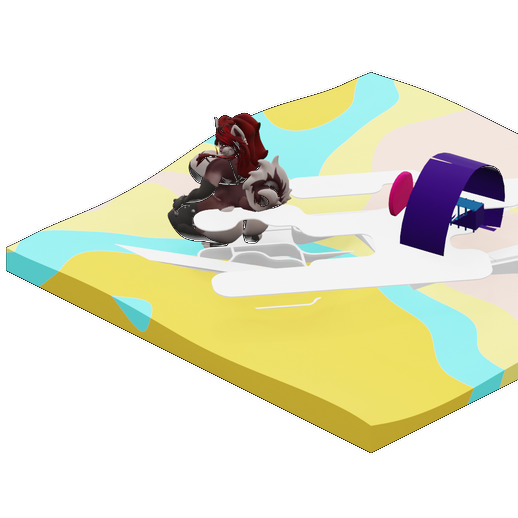} & \includegraphics[width=0.180\textwidth]{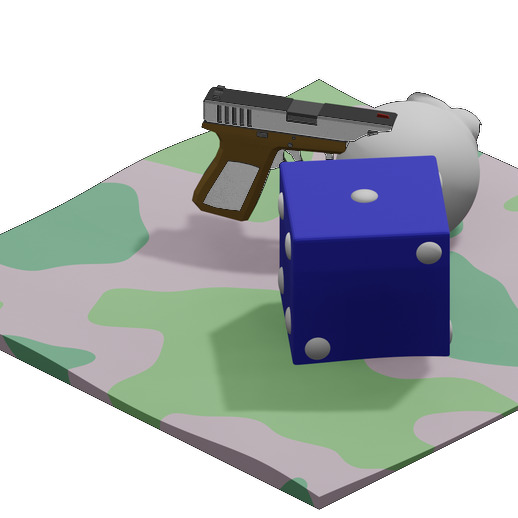} & \includegraphics[width=0.180\textwidth]{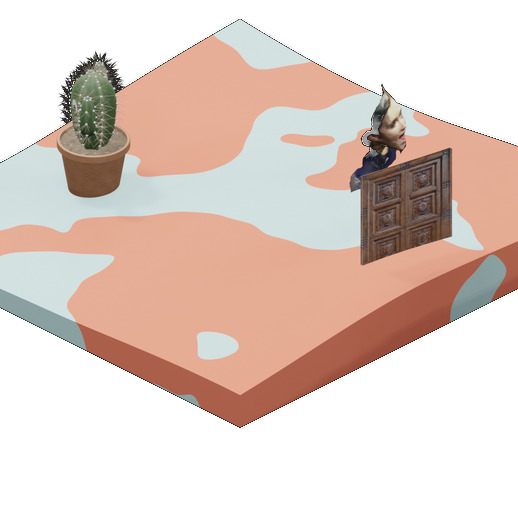} \\
\end{tabular}
\caption{Random selection of samples generated by the dataset engine proposed in Section 5.}%
\label{fig:supp-dataset-samples}
\end{figure*}

\paragraph{Dataset samples.}
We present a random selection of rendered 3D samples generated by our data engine in \cref{fig:supp-dataset-samples}.
We refer to these as the ``Conditioning Image'' in Figure 5.
For visualization purposes, we removed the black background.

\clearpage

\end{document}